\pdfoutput=1

\documentclass{article}

\usepackage{arxiv}

\usepackage[utf8]{inputenc} 
\usepackage{hyperref}       
\usepackage{url}            
\usepackage{booktabs}       
\usepackage{amsfonts}       
\usepackage{nicefrac}       
\usepackage{microtype}      
\usepackage[title]{appendix}
\usepackage{graphicx}

\usepackage{array}
\usepackage{multirow}
\usepackage{amsmath}

\usepackage[most]{tcolorbox}
\tcbuselibrary{listings}

\usepackage[markup=underlined,commandnameprefix=ifneeded]{changes}

\newtcblisting{promptbox}{
    breakable,
    listing only,
    colback=gray!8,        
    colframe=gray!40,      
    arc=4pt,               
    boxrule=0.5pt,         
    left=1em,              
    right=1em,             
    top=0.5em,             
    bottom=0.5em,          
    listing options={
        basicstyle=\ttfamily\small,
        breaklines=true,
        numbers=none
    }
}

\usepackage{doi}
\usepackage{etoolbox}
\usepackage[natbibapa]{apacite}
\AtBeginDocument{}
\renewcommand{\APACrefnote}[1]{}
\newtoggle{bibdoi}
\newtoggle{biburl}
\makeatletter
\newsavebox{\bib@url}
\newsavebox{\bib@doi}

\undef{\APACrefURL}
\undef{\endAPACrefURL}
\undef{\APACrefDOI}
\undef{\endAPACrefDOI}

\newenvironment{APACrefURL}
  {\global\toggletrue{biburl}\lrbox\bib@url}
  {\endlrbox}

\newenvironment{APACrefDOI}
  {\global\toggletrue{bibdoi}\lrbox\bib@doi}
  {\endlrbox}

\newcommand{\printinfo}{
  \iftoggle{bibdoi}{\usebox{\bib@doi}}{\usebox{\bib@url}}
  \togglefalse{bibdoi}
}

\AtBeginEnvironment{thebibliography}{
\pretocmd{\PrintBackRefs}{%
  \iftoggle{bibdoi}
    {\iftoggle{biburl}{\unskip\unskip}{}\usebox{\bib@doi}}
    {\iftoggle{biburl}{Retrieved from \usebox{\bib@url}}}{}
  \togglefalse{bibdoi}\togglefalse{biburl}%
}{}{}}
\makeatother

\title{Wildfire-SSR: Geolocation-Aware Multimodal Fusion for Structure-Level Wildfire Risk Assessment}

\date{} 					

\renewcommand{\headeright}{}
\renewcommand{\undertitle}{}
\renewcommand{\shorttitle}{}

\hypersetup{
pdftitle={},
pdfsubject={},
}

\begin{document}
\maketitle

\begin{center}
{\Large
Yu-Hsuan Ho\textsuperscript{a,*},
Yiming Xiao\textsuperscript{a},
Hao-En Lai\textsuperscript{b},
Ali Mostafavi\textsuperscript{a,c},
Samuel D. Brody\textsuperscript{c}
\par}

\bigskip
\textsuperscript{a} Urban Resilience.AI Lab, Zachry Department of Civil and Environmental Engineering,\\ Texas A\&M University, College Station, TX\\
\textsuperscript{b} Artie McFerrin Department of Chemical Engineering,\\ Texas A\&M University, College Station, TX\\
\textsuperscript{c} Institute for Disaster Resilient Texas, Texas A\&M  University, Houston, TX\\
\vspace{6pt}
\textsuperscript{*} corresponding author, email: yuhsuanho@tamu.edu
\\
\end{center}
\bigskip
\begin{abstract}
Structure-level wildfire vulnerability assessment is essential for reducing losses in wildland-urban interface communities, yet existing approaches struggle to combine parcel-scale ignition pathways with community-scale geographic coverage. This paper argues that accurate and scalable assessment requires a geolocation-aware multimodal framework that integrates macro-scale satellite context with micro-scale street-view evidence. To operationalize that argument, the study introduces Wildfire-SSR (Satellite and Street-view for wildfire Risk), an automated pipeline that uses a Vision-Language Model to extract wildfire-relevant building and site attributes from pre-event street-view panoramas, converts those observations into structured semantic text, and spatially aligns them with satellite image features through a decoupled cross-attention fusion decoder. A frozen satellite-pretrained DINOv3 encoder captures overhead visual context, while alternative geolocation embedding strategies are evaluated to anchor the text-derived features within each satellite patch. The framework is tested on approximately 37,000 structures affected by the 2025 Eaton and Palisades fires in Los Angeles, using wildfire risk labels derived from relative post-event damage patterns to better reflect inherent structural vulnerability. Results show that Rotary Positional Embedding provides the strongest spatial grounding in the real-world setting, yielding a building localization F1 score of 0.6735 and a harmonic mean F1 score of 0.2476 for five-class pixel-level risk classification, while outperforming both unimodal baselines and an unanchored multimodal configuration. The model also exhibits a conservative bias in higher-risk classes, reducing the likelihood of underestimating severe vulnerability, and completes end-to-end assessment of the study area in 37 minutes on a single GPU. Scientifically, the study advances multimodal remote sensing analytics by reframing cross-view fusion as localized semantic alignment. Practically, it provides a scalable pipeline for mitigation prioritization, resilience planning, and insurance-oriented wildfire risk assessment.
\end{abstract}



\section{Introduction}
\label{sec:introduction}
Wildfire losses in wildland-urban interface (WUI) communities are shaped not only by the intensity of fire behavior at the landscape scale, but also by the ignition susceptibility of individual structures \citep{calkin2014risk}. This makes structure-level vulnerability assessment a central problem for resilience planning, because mitigation decisions are implemented at the parcel and building level even when hazards unfold across regions. Yet the prevailing alternatives remain unsatisfactory: manual inspections provide rich detail but are too labor-intensive to scale, while automated overhead analyses can cover large areas but overlook the facade-level conditions that often determine whether a structure ignites. As wildfire exposure expands across complex urban-wildland settings, the ability to detect both community-scale exposure patterns and building-scale vulnerabilities becomes increasingly important. This paper therefore tackles a problem of high scientific and societal consequence: how to produce structure-specific wildfire intelligence without sacrificing geographic coverage.

Remote sensing has long served as the foundation for scalable disaster risk analytics \citep{platt2014wildfire, ahmed2018remote}, efficiently mapping macro-level features such as building footprints, canopy cover, and regional vegetation continuity \citep{wu2025semantic, xiao2025damagecat, ho2025multimodal, hu2024high, kaur2023large}. However, top-down vantage points cannot capture the ground-level drivers of structure ignition such as combustible siding, unenclosed eaves, debris accumulation, or vegetation in direct contact with exterior walls \citep{dossi2023relationships, papathoma2022wildfire, syphard2017importance}. Street-view imagery recovers these facade-level attributes \citep{xia2025integrating, ho2024elev, xiao2025recov, esparza2025automated, li2026property}. However, it lacks the landscape and terrain context that satellite basemaps provide, so a comprehensive assessment naturally requires integrating both modalities \citep{esparza2025graphfire}. 

Existing cross-view fusion paradigms assume a one-to-one mapping between a single satellite patch and a single street-view image \citep{hu2018cvm, shi2022beyond, wang2021each, liu2025gair}, which is architecturally and computationally inefficient at community scale. Because a high-resolution satellite image patch covers a broad geographic extent containing multiple structures, it must instead be paired with a corresponding batch of street-view images to accurately evaluate the entire scene. Processing a large overhead basemap alongside multiple high-resolution street-view images through a joint visual architecture rapidly exceeds practical computing resource limits. Instead, real-world success requires a deployment-centric approach that proactively aligns with practical constraints from the outset \citep{liu2025towards}. Vision-Language Models (VLMs) offer a practical alternative by compressing high-resolution street-view imagery into lightweight structured text \citep{xiao2025recov, esparza2025automated}. These models have demonstrated remarkable proficiency in the zero-shot and few-shot extraction of structured semantic information from complex structural and urban environments \citep{ho2025integrated, hu2025rsgpt, kuckreja2024geochat}. This approach preserves critical semantic attributes, such as roof material, defensible space, and debris presence, while drastically reducing the computational burden. Text-based compression resolves scalability, but introduces a distinct challenge of spatial alignment. Current methods rely on vision-based cross-view localization \citep{hu2018cvm, shi2022beyond, wang2021each, zhu2021vigor} while underutilizing a readily available signal: satellite imagery is georeferenced and street-view captures carry precise coordinates. 

To capture geographic context, researchers have increasingly explored explicit geolocation embedding methods that project raw coordinates into feature spaces. Early approaches relied on concatenating explicit geographic data with visual features \citep{tang2015improving, christie2018functional}. More recent studies have advanced beyond discrete metadata by learning continuous semantic embeddings directly from GPS coordinates \citep{yin2019gps2vec, vivanco2023geoclip}. This continuous modeling approach has also been adapted specifically for remote sensing foundation models \citep{klemmer2025satclip, liu2025gair}. While these existing methods successfully and effectively bridge geospatial embeddings with dense visual architectures, they are fundamentally designed to learn global location embeddings suitable for broad, macro-level area studies with simplified one-to-one image-coordinate alignment. Assessing fine-grained, community-scale vulnerabilities instead requires a local, many-to-one architecture capable of spatially anchoring multiple street-level observations within a single high-resolution satellite basemap. The adaptable positional encodings developed for Vision Transformers (ViT), including fixed sinusoidal encodings \citep{vaswani2017attention}, learnable absolute encodings \citep{dosovitskiy2020image}, relative encodings \citep{liu2021swin}, Rotary Position Embedding (RoPE) \citep{su2024roformer}, and Random Fourier Features (RFF) \citep{tancik2020fourier}, provide a principled mechanism to inject such fine-grained spatial awareness, but their use for geolocation-aware fusion of satellite imagery and multiple street-view texts remains largely unexplored.

\begin{figure}[!ht]
    \centering
    \includegraphics[width=\textwidth]{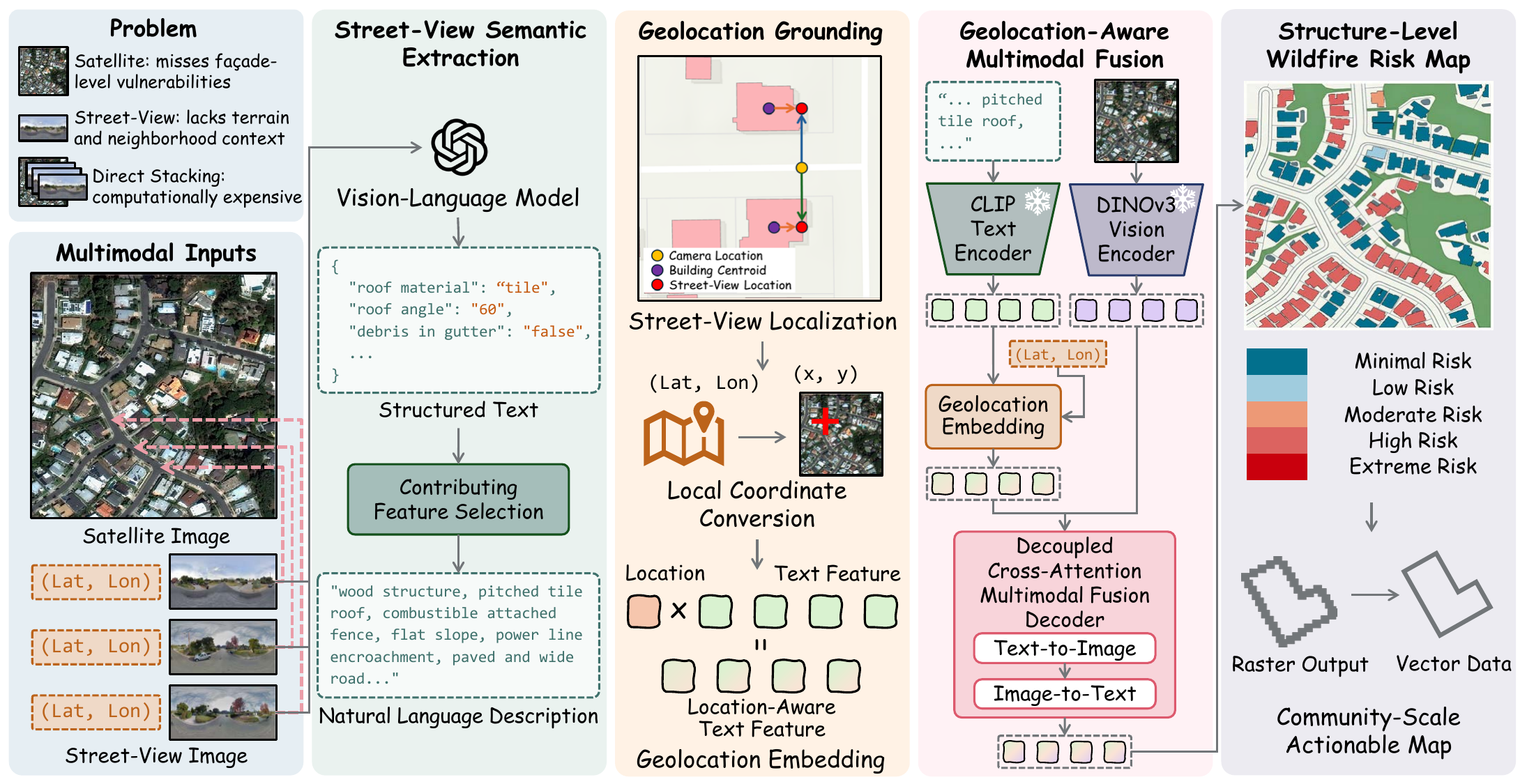}
    \caption{Overview of the proposed geolocation-aware multimodal framework for structure-level wildfire risk assessment. The pipeline addresses the limitations of isolated unimodal approaches by integrating macro-scale satellite imagery with micro-scale street-view data. A Vision-Language Model first extracts semantic building vulnerabilities from street-level images into structured text descriptions. To bridge the spatial disconnect between modalities, the geographical coordinates of the street-view captures are mapped to local relative coordinates and injected into CLIP-encoded text embeddings. A decoupled cross-attention multimodal fusion decoder then aligns these geolocation-aware text features with overhead visual features extracted via DINOv3. The proposed model processes these fused representations into pixel-wise risk labels, which are then transformed into a detailed, actionable community-scale wildfire risk map.}
    \label{fig:conceptual_figure}
\end{figure}

Taken together, existing wildfire vulnerability studies remain fragmented across scale and modality. Neither macro-scale remote sensing nor micro-scale street imagery is sufficient on its own, and na\"ive multimodal fusion does not scale. This paper departs from that literature by treating structure-level wildfire assessment as a many-to-one spatial alignment problem, addressing not only a data-fusion gap but a scalability gap that has limited the operational use of multimodal wildfire analytics. The rationale for the proposed design is that wildfire vulnerability is jointly produced by parcel-specific ignition conditions and surrounding environmental context, so the model must represent both. To this end, we propose a geolocation-aware multimodal framework for structure-level wildfire risk assessment. A VLM first extracts detailed building attributes from pre-event street-view imagery into structured text, while a satellite-pretrained DINOv3 \citep{simeoni2025dinov3} vision foundation model extracts macro-level features from the corresponding satellite patch. The framework then bridges the text-imagery spatial disconnect by mapping the latitude and longitude of each street-view capture to relative pixel coordinates within the satellite patch and injecting this signal into CLIP-encoded \citep{radford2021learning} text features through three alternative strategies: RoPE \citep{su2024roformer} for relative dependencies, RFF \citep{tancik2020fourier} for absolute high-dimensional embeddings, and a Spatial Grid Mapping that places text features into a 2D spatial tensor at their grid coordinates. A cross-attention fusion decoder finally aligns these spatially aware text features with overhead visual features to produce structure-level vulnerability predictions. Figure \ref{fig:conceptual_figure} illustrates the proposed workflow.

Against this backdrop, the study pursues three linked objectives: (i) to develop a computationally tractable framework that integrates pre-event satellite imagery with street-view-derived structural semantics for structure-level wildfire vulnerability assessment; (ii) to determine how ground-level information should be spatially anchored within overhead scenes through alternative geolocation embedding strategies; and (iii) to evaluate whether this multimodal, geolocation-aware design improves prediction quality and operational scalability in real WUI fire contexts. By testing the framework on the Eaton and Palisades fires and pairing ablation analysis with diagnostic experiments, the paper moves beyond proof-of-concept model building toward an evidence-based assessment of when and why multimodal spatial alignment matters.

\section{Methodology}
\label{sec:methodology}
\begin{figure}[ht]
    \centering
    \includegraphics[width=\textwidth]{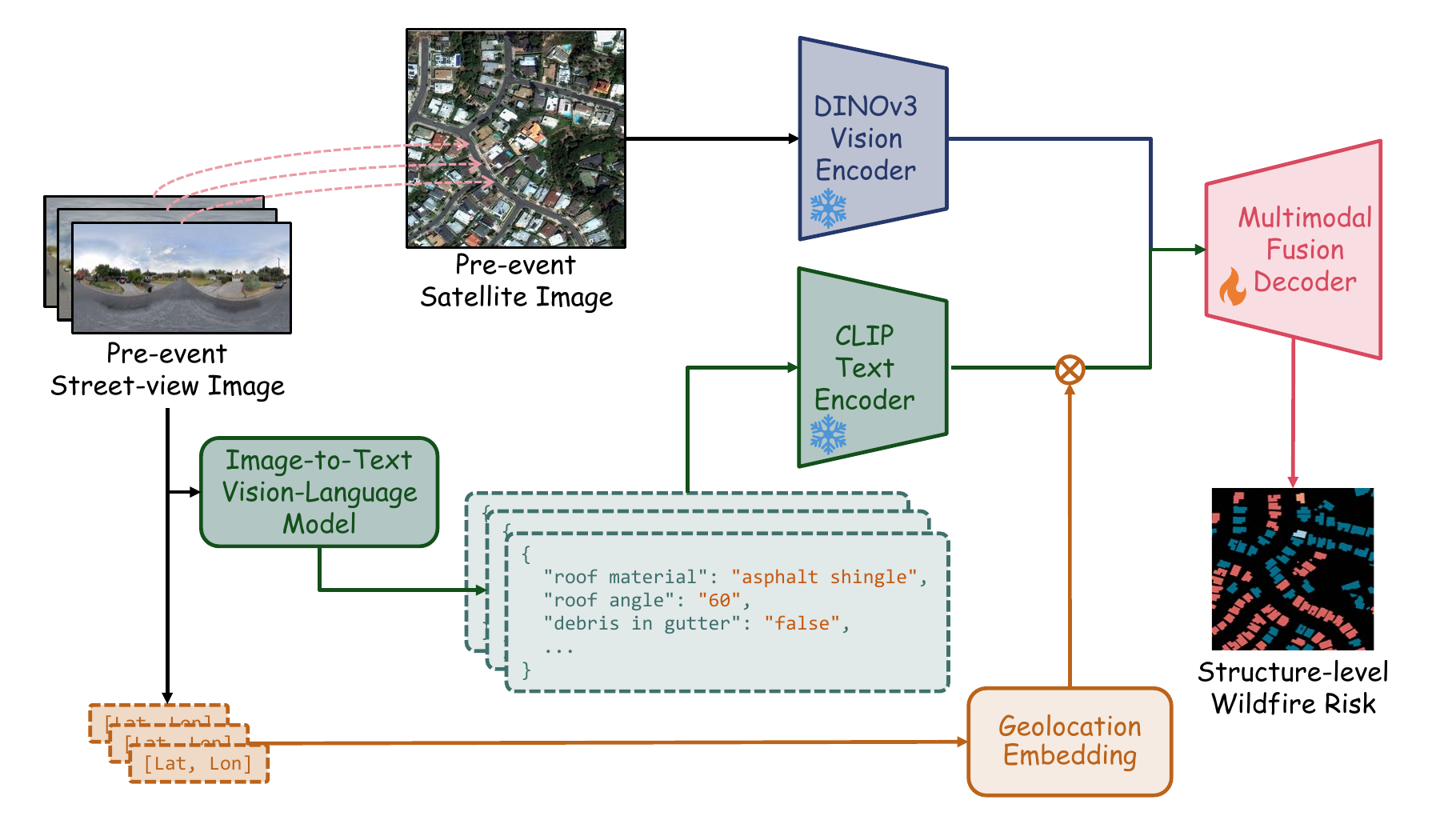}
    \caption{Overview of the model architecture. Ground-level street-view images are first processed by a Vision-Language Model to extract structured semantic attributes, which are combined with their geographic coordinates to form spatially anchored text embeddings. A multimodal fusion decoder then integrates these location-aware micro-level features with macro-level visual features extracted from pre-event satellite imagery via a frozen satellite-pretrained vision encoder to generate precise, structure-level wildfire risk maps.}
    \label{fig:network_architecture}
\end{figure}

The proposed framework introduces a novel deep-learning architecture designed for the geolocation-aware multimodal fusion of overhead satellite imagery and ground-level street-view data. To manage computational overhead and extract structured building attributes, the raw street-view images are first converted into semantic text descriptions using a Vision-Language Model prior to multimodal fusion. To achieve highly accurate, structure-level wildfire vulnerability intelligence, the system overcomes the computational and spatial alignment bottlenecks of traditional multimodal models. It leverages pre-trained, frozen vision and language foundation models to extract robust macro-level visual features and micro-level semantic embeddings. The core of the framework is a custom, spatially-aware fusion decoder that aligns these disparate modalities by capturing explicit spatial dependencies before outputting precise building footprint and vulnerability maps. Figure \ref{fig:network_architecture} illustrates the proposed network architecture. 

\subsection{Network Architecture}
The architecture is structured as a dual-encoder, single-decoder UNet-style network. It processes two distinct input streams: pre-event satellite image patches and a corresponding batch of VLM-derived text descriptions. The raw geographical coordinates of these street-view captures are first converted into localized, normalized coordinates prior to ingestion. This conversion maps the exact geographical location of the text description to a relative 2D pixel coordinate $(x, y)$ bounded between $-1$ and $1$ within the paired satellite image patch, indicating exactly where the ground-level view sits within the specific overhead perspective. The network consists of four primary components: a Vision Encoder, a Language Encoder, a Multimodal Fusion Decoder, and Dual Segmentation Heads.

The visual pathway utilizes a DINOv3 \citep{simeoni2025dinov3} Vision Transformer (ViT-L/16) pre-trained on SAT-493M \citep{simeoni2025dinov3}, a large satellite dataset, as its backbone. To retain the generalized feature representations learned during its extensive pre-training, the DINOv3 backbone is completely frozen. The backbone is wrapped in a custom adapter equipped with interaction blocks to extract hierarchical, multi-scale visual features from specific intermediate layers of the ViT, providing the decoder with a rich pyramid of macro-level contextual features.

The textual pathway relies on a pre-trained, frozen CLIP \citep{radford2021learning} text encoder (ViT-B/16). It is designed to handle a variable number of street-view text descriptions per satellite tile. The module tokenizes and encodes the raw text, passing the outputs through a linear projection layer to ensure the semantic feature dimensions match the embedding dimension of the visual backbone. To accommodate ragged inputs within a batch, the encoder constructs padded feature tensors and Boolean attention masks. Specifically, the encoder dynamically finds the maximum number of texts in a batch and pads the shorter sequences with zeros, marking them with an ignore mask. Crucially, this module processes and forwards the normalized 2D geographical coordinates associated with each text feature for downstream spatial alignment.

\begin{figure}[ht]
    \centering
    \includegraphics[width=0.5\textwidth]{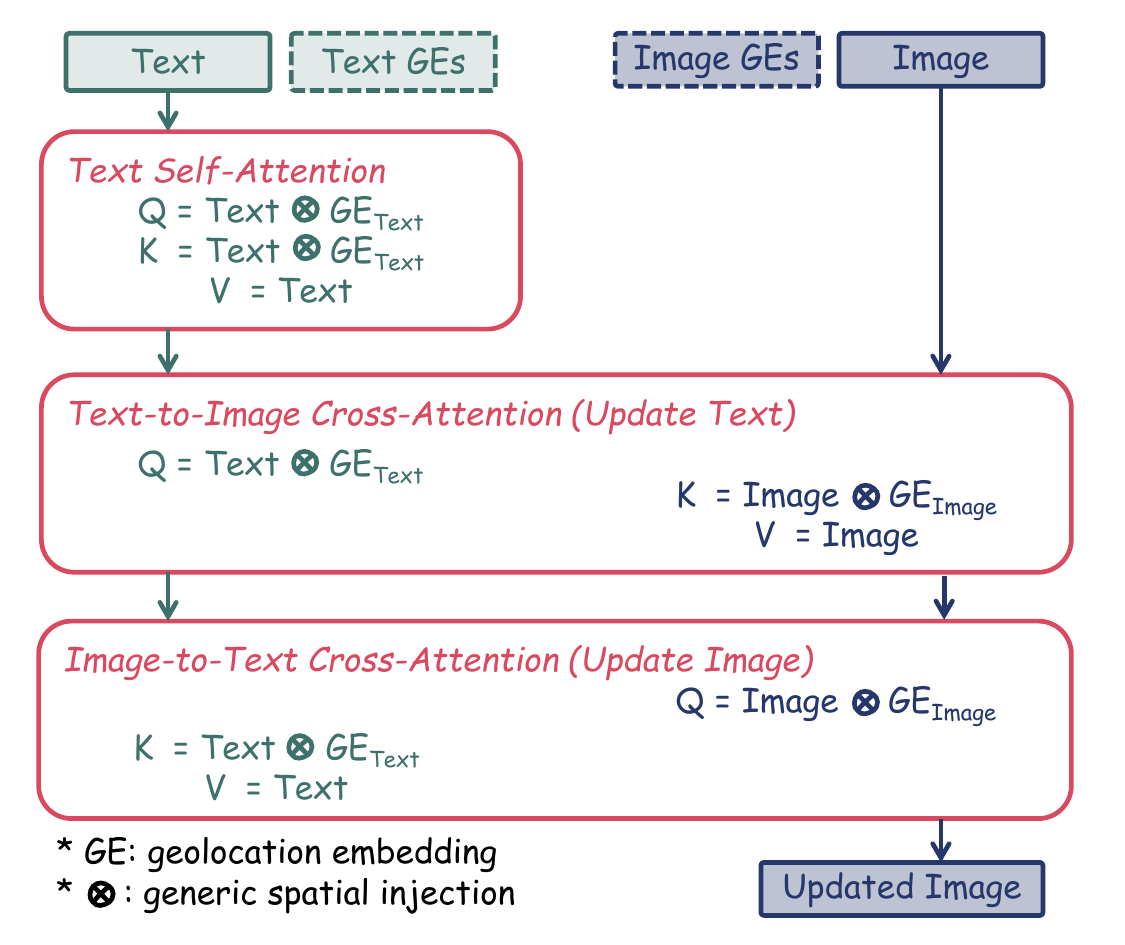}
    \caption{Structure of the two-way cross-attention block utilizing a decoupled attention mechanism. Geolocation embeddings (GEs) are injected specifically into the queries ($Q$) and keys ($K$) to facilitate spatial routing, while values ($V$) remain purely semantic to preserve micro-level building attributes. The block consists of three sequential steps: sparse text self-attention to capture geometric dependencies between structural, sparse-to-dense cross-attention to update textual context with overhead visual data, and dense-to-sparse cross-attention to infuse the macro-level visual map with ground-sourced semantic details.}
    \label{fig:two_way_cross_attention}
\end{figure}

The decoder is the central fusion engine of the architecture. It operates across multiple stages, gradually upsampling the spatial resolution while integrating the multi-scale visual features via skip connections. At each stage, a two-way cross attention block performs the multimodal alignment. 
Figure \ref{fig:two_way_cross_attention} illustrates the two-way cross attention block. To spatially ground the textual features, this block dynamically injects geolocation embeddings, derived from the normalized spatial coordinates, directly into the attention layers. (The specific positional embedding strategies explored to achieve this spatial awareness are detailed in Section \ref{sec: Geolocation Embeddings}.) This is achieved using a decoupled attention mechanism inspired by the two-way transformer architecture of the Segment Anything Model (SAM) \citep{kirillov2023segment}. Unlike standard transformer attention where positional data is baked into the input sequence before projection, decoupled attention mathematically separates spatial and semantic information. The positional embeddings are injected strictly into the queries ($Q$) and keys ($K$) to dictate spatial routing, guiding exactly where the model should look and how features geometrically relate, while the values ($V$) remain purely semantic. This ensures that the rich visual and textual data passed forward to the segmentation heads is highly spatially aware but untainted by coordinate transformations. The multimodal alignment occurs through a sequence of three such decoupled attention mechanisms:

\begin{itemize}
    \item Sparse Text Self-Attention (Text-to-Text): The localized text queries and keys interact with the injected spatial embeddings, allowing the text features to attend to one another and build a cohesive understanding of the ground-level scene based on their geometric relationships.

    \item Sparse-to-Dense Cross-Attention (Text-to-Image): The spatially-aware text queries attend to the dense image features, updating the textual context based on the surrounding overhead visual data.

    \item Dense-to-Sparse Cross-Attention (Image-to-Text): The dense image queries attend to the refined, spatially-anchored text features, effectively infusing the macro-level visual map with micro-level, ground-sourced semantic details.
\end{itemize}

\subsection{Geolocation Embeddings}
\label{sec: Geolocation Embeddings}
To ensure that the micro-level textual attributes are accurately anchored to the macro-level satellite basemap, the localized and normalized 2D geographical coordinates associated with each street-view semantic representation must be transformed into robust, high-dimensional representations. Because raw $x$ and $y$ coordinates lack the dimensionality required to effectively influence dense visual attention mechanisms, the framework evaluates three geolocation embedding strategies. Each method bridges the spatial disconnect by injecting geographical awareness into the multimodal fusion pipeline.

\subsubsection{Rotary Positional Encoding (RoPE)}
The Rotary Positional Encoding \citep{su2024roformer} strategy rotates the semantic representations based on their relative spatial distances rather than adding static positional vectors to the input features. In this approach, the 2D normalized coordinates are translated into a spectrum of varying angular frequencies across the feature dimensions, allowing the model to capture both fine- and coarse-grained spatial relationships. Specifically, for a normalized coordinate vector $\mathbf{c} \in [-1, 1]^2$ and a sequence of exponential periods $\mathbf{P}$, the angular frequency matrix is defined as $\Theta = \frac{2\pi \mathbf{c}}{\mathbf{P}}$, where the $x$ and $y$ components of $\mathbf{c}$ are applied to separate halves of the feature channels. These angles are then converted into sine and cosine transformations, which are applied to the query and key feature vectors ($\mathbf{x}$) within the decoupled attention block via a 2D rotational shift:
$$\mathbf{x}_{rotated} = \mathbf{x} \odot \cos(\Theta) + \mathbf{x}_{half} \odot \sin(\Theta)$$
where $\mathbf{x}_{half} = [-\mathbf{x}_{d/2:}, \mathbf{x}_{:d/2}]$ represents the swapped and partially negated feature dimensions. By modulating the attention routing through rotational mathematics, RoPE explicitly captures the relative distance dependencies between different building attributes without distorting the underlying semantic values.

\subsubsection{Random Fourier Features (RFF)}
The Random Fourier Features \citep{tancik2020fourier} strategy projects the low-dimensional geographic coordinates into a high-dimensional continuous space to capture fine-grained, high-frequency spatial details. The normalized coordinate vectors $\mathbf{c} \in [-1, 1]^2$, are first multiplied by a randomly initialized Gaussian weight matrix $\mathbf{W} \sim \mathcal{N}(0, \sigma^2)$. The resulting high-dimensional projections are passed through sine and cosine trigonometric functions and concatenated to form absolute positional embeddings:
$$PE_{RFF}(\mathbf{c}) = \left[ \sin(2\pi \mathbf{c} \mathbf{W}), \cos(2\pi \mathbf{c} \mathbf{W}) \right]$$
Unlike the dynamic rotation of RoPE, these static spatial embeddings are added directly to the query and key tensors prior to their linear projection within the decoupled attention mechanism.

\subsubsection{Spatial Grid Mapping}
Instead of relying on attention mechanisms to implicitly learn spatial alignments, the spatial grid mapping method explicitly preserves the physical geographic arrangement by building a 2D feature map. The normalized coordinate vectors $\mathbf{c} \in [-1, 1]^2$ are scaled and discretized into exact pixel indices $(x_{grid}, y_{grid})$ that match the resolution of the satellite image patch grid:
$$(x_{grid}, y_{grid}) = \left( \min \left( \left\lfloor \frac{x + 1}{2} \cdot \frac{S}{P} \right\rfloor, S - 1 \right), \min \left( \left\lfloor \frac{y + 1}{2} \cdot \frac{S}{P} \right\rfloor, S - 1 \right) \right)$$
where $S$ is the image size and $P$ is the model's patch size. The high-dimensional CLIP-derived semantic features are then inserted directly into an empty spatial tensor at their corresponding calculated grid locations. This approach maps the isolated street-view texts into a physical layout, creating a semantic tensor that can be directly processed by a 2D Transformer encoder and concatenated with the visual features.

\subsection{Street-View Semantic Feature Extraction}
\label{sec: sv_extraction}
To derive localized building attributes from ground-level perspectives, the framework employs a Vision-Language Model (VLM), specifically GPT-4o \citep{gpt4o}, to process panoramic street-view imagery. The VLM is prompted to systematically evaluate the images and extract a comprehensive set of predefined structural and environmental characteristics, such as roof material, siding type, and vegetation-structure contact. The complete list of these predefined characteristics, the specific prompts utilized, and example outputs are detailed in Appendix \ref{app:vlm_prompt} and \ref{app:example_outputs}. The model outputs this micro-level data as structured, categorical key-value pairs, which are saved in JSON format. To maintain data integrity and prevent hallucination, the VLM is instructed to return the value ``unsure'' for any attribute that is visually obscured or undefinable from the given viewpoint.

Generating highly efficient and accurate text embeddings requires isolating the most predictive building attributes while minimizing noise. To refine the extracted street-view semantic texts, recursive feature elimination with cross-validation is implemented using the training set. This automated feature selection process utilizes an XGBoost \citep{chen2016xgboost} classifier as the base estimator, iteratively evaluating feature importance and dropping the single least important variable in each round. The optimization process is driven by the harmonic mean F1 score and rigorously validated using 5-fold stratified cross-validation, which guarantees that the original imbalanced class distribution of the dependent variable is proportionally preserved across every training and testing fold to prevent skewed evaluation metrics. Rather than relying on a manually defined threshold, the optimal number of independent variables is automatically determined by tracking the model's performance iteratively and permanently selecting the exact feature subset size that yields the highest cross-validated peak score. This elimination strategy successfully identifies and removes 9 non-contributing features, a 36\% reduction from an initial 25 features down to 16, without any degradation in overall model performance. The full list of features is detailed in Appendix \ref{app:feature_dictionary}.

Finally, the refined, structured data must be transformed back into a format suitable for the CLIP text encoder. A rule-based natural language generation pipeline performs a structured-to-unstructured conversion, translating the categorical feature dictionaries into coherent, comma-separated descriptive captions. If an attribute is successfully identified, the pipeline generates a precise phrase (e.g.,``wood structure''); if the VLM marked the attribute as ``unsure'', the logic defaults to a generalized term (e.g., ``building''). Regarding the identification of complex risk states, descriptive terms are omitted entirely if no risk is identified. For instance, if the VLM outputs ``no power lines'', the pipeline does not explicitly describe this absence; rather, it excludes the attribute from the caption altogether. This preprocessing ensures that the final text inputs provided to the language encoder remain semantically dense and free of unnecessary noise. Examples of this conversion are detailed in Appendix \ref{app:natural_language}.

\subsection{Street-View Localization}
To accurately anchor the ground-level semantic text features to the overhead satellite basemap, the precise geographic coordinates of the observed building facades must be calculated. Because raw street-view metadata only provides the coordinate of the vehicle or camera in the center of the street, the framework implements a geometric localization algorithm to project these coordinates onto the nearest adjacent structures. Simply linking the semantic text to the centroid of the nearest building is insufficient, as it results in a critical loss of view-angle information. A single structure is frequently captured by multiple street-view images from distinct vantage points; therefore, projecting the coordinates geometrically ensures that the exact observed facade and directional perspective are preserved within the spatial embedding.

\begin{figure}[ht]
    \centering
    \includegraphics[width=0.5\textwidth]{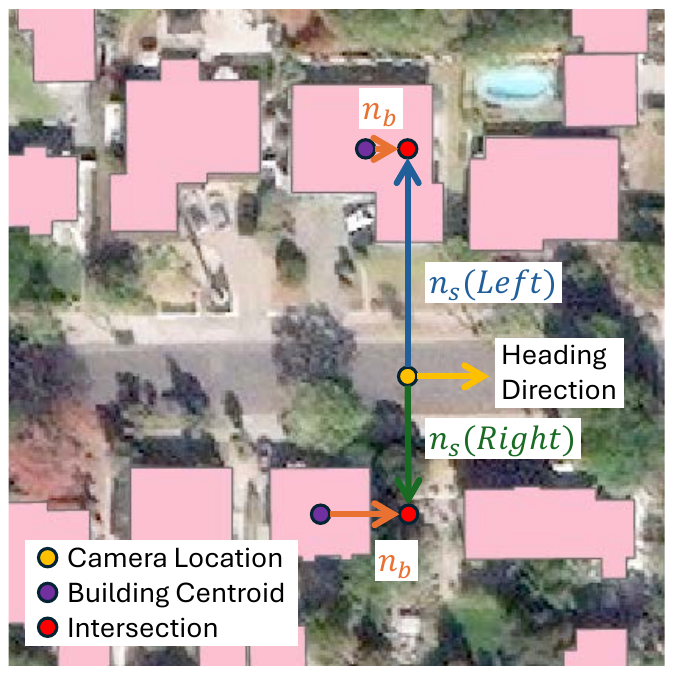}
    \caption{Schematic representation of the geometric localization algorithm. To preserve view-angle perspective, orthogonal normal vectors ($\mathbf{n}_s$) are cast from the street-view camera's heading direction. The intersection of these street vectors with the heading vectors ($\mathbf{n}_b$) originating from adjacent building centroids determines the precise spatial anchor point on the observed facade.}
    \label{fig:street-view_localization}
\end{figure}

The algorithm processes each street-view capture by first defining a heading direction vector based on the camera's recorded orientation. It then identifies the closest structures on both the left and right sides of the street. Once the target building is identified, a normal vector is cast orthogonally from the camera's heading direction, clockwise for buildings on the right and counterclockwise for buildings on the left, as illustrated in Figure \ref{fig:street-view_localization}. Using spherical trigonometry principles, the exact geospatial intersection point between the orthogonal street vector from the camera and the heading direction vector from the building center is calculated by representing their paths as great circles. Let $\mathbf{n}_s$ and $\mathbf{n}_b$ represent the 3D normal vectors defining the planes of the street's great circle and the building's great circle, respectively, derived from their spherical coordinates and bearings. The Cartesian intersection vector $\mathbf{v}_{int}$ is defined by their normalized cross product:
$$\mathbf{v}_{int} = \pm \frac{\mathbf{n}_s \times \mathbf{n}_b}{\| \mathbf{n}_s \times \mathbf{n}_b \|}$$
where the sign is selected to yield the intersection point closest to the original coordinates. This Cartesian vector $(x, y, z)$ is then converted back into the localized latitude ($\phi_{int}$) and longitude ($\lambda_{int}$) via:$$\phi_{int} = \arcsin(z), \quad \lambda_{int} = \arctan2(y, x)$$

To prevent erroneous projections in sparse environments, where empty lots, missing structures, or deep setbacks might cause the algorithm to inadvertently project the intersection onto a distant, unrelated building, the algorithm enforces a maximum projection distance constraint. According to the City of Los Angeles Complete Streets Design Guide \citep{lacity_completestreets}, residential Rights-of-Way (ROW) generally range from roughly 8 meters on narrow hillside streets to 20 meters on standard collector streets. When standard front yard setbacks of 4.5 to 9 meters are added \citep{lamc_article2_2025}, the total street corridor spans approximately 17 to 38 meters. Consequently, the expected projection distance from the camera's origin at the street center to an adjacent building facade is roughly 8.5 to 19 meters. Therefore, a 25-meter threshold provides an adequate boundary to capture true adjacent facades while filtering out false positives. If the calculated intersection distance exceeds this limit, the localized point is automatically capped at the 25-meter mark along the orthogonal vector. The resulting localized coordinates, representing the exact observed facade location, are then normalized and fed into the multimodal decoder's spatial embedding mechanisms.

\subsection{Loss Function}
\label{sec: Loss Function}

Because the proposed framework simultaneously performs binary building localization and multi-class wildfire risk assessment, it is trained using a multi-task composite loss function. Both tasks inherently suffer from class imbalances. To mitigate this issue, the framework combines pixel-level entropy losses with structural overlap optimizations. Background pixels are ignored across all loss calculations.

\subsubsection{Localization Loss}
To optimize the building footprint extraction, the model employs a weighted combination of standard Cross-Entropy (CE) loss and Lovász-Softmax loss \citep{berman2018lovasz}. While CE loss evaluates the independent probability distribution of each pixel, the Lovász-Softmax loss directly acts as a differentiable surrogate for the Intersection-over-Union (IoU) metric, effectively handling spatial scale imbalances between large and small structures. The Lovász-Softmax loss is mathematically defined as
$$\mathcal{L}_{Lovasz} = \frac{1}{|C|} \sum_{c \in C} \sum_{i=1}^{P} e_i(c) g_i(c)$$
where $e_i(c) = |y_i(c) - p_i(c)|$ represents the absolute pixel errors for class $c$ sorted in descending order, and $g_i(c)$ is the Lovász extension gradient that calculates the discrete Jaccard index (IoU) penalty over the sorted sequence. The localization loss is formulated as
$$\mathcal{L}_{loc} = \lambda_{ce}^{loc} \mathcal{L}_{CE} + \lambda_{lov}^{loc} \mathcal{L}_{Lovasz}$$

\subsubsection{Classification Loss}
For the wildfire risk classification task, the model must account for the ordinal nature of risk severity. Misclassifying an extreme-risk structure with as low risk is far more egregious than misclassifying it as high risk. Therefore, the classification loss replaces standard cross-entropy with a Distance-Weighted Cross-Entropy (WCE) loss, combined again with the Lovász-Softmax loss. 

To simultaneously counteract the dataset imbalance, class-specific penalty weights are integrated into the Cross-Entropy calculation. The frequencies are calculated at the discrete building object level. The weight $w_c$ for a specific risk class $c$ is defined by its inverse frequency within the training dataset:
$$w_c = \frac{N}{N_c}$$
where $N$ is the total number of building instances and $N_c$ is the total number of buildings assigned to class $c$. For any single valid pixel, this balancing factor is applied directly to the foundational Cross-Entropy component ($\mathcal{L}_{CE}$) to amplify the learning gradient for minority classes:
$$\mathcal{L}_{CE} = - \sum_{c=0}^{C-1} w_c \left( y_{c} \cdot \log(p_{c}) \right)$$
where $y_{c} \in \{0,1\}$ is the one-hot encoded ground truth for class $c$, and $p_{c}$ is the predicted probability for that class.

The WCE then penalizes predictions based on the ordinal distance between the predicted and true risk classes, modulated by predefined class weights and a distance scalar parameter $\alpha=0.5$. Specifically, it calculates the expected continuous class value using the softmax probabilities and adds a Mean Squared Error (MSE) penalty to the standard Cross-Entropy ($\mathcal{L}_{CE}$):
$$\mathcal{L}_{WCE} = \mathcal{L}_{CE} + \alpha \left( \sum_{c=0}^{C-1} p_c \cdot c - y \right)^2$$
where $p_c$ is the predicted probability for class $c$, and $y$ is the ground truth class integer. While the WCE loss strictly penalizes these ordinal severity errors, the accompanying Lovász-Softmax loss operates independently of ordinal distance, focusing on maximizing the structural IoU for each risk class to ensure sharp spatial boundaries. The total classification loss is defined as
$$\mathcal{L}_{clf} = \lambda_{ce}^{clf} \mathcal{L}_{WCE} + \lambda_{lov}^{clf} \mathcal{L}_{Lovasz}$$

The final objective function minimizes the sum of both task losses $\mathcal{L}_{total} = \mathcal{L}_{loc} + \mathcal{L}_{clf}$.

\section{Experimental Setup}
The proposed framework was evaluated using the 2025 Eaton and Palisades fires in Los Angeles, California, as two real-world wildland–urban interface case studies. The experimental setup integrates pre-event satellite imagery, VLM-derived pre-event street-view semantic texts, building footprints, and post-event CAL FIRE damage inspection records to construct a multimodal benchmark for structure-level wildfire vulnerability assessment. This section describes the study areas and data sources, the derivation of neighborhood-relative vulnerability labels from observed structural damage, the spatial partitioning and augmentation procedures used to prevent geographic data leakage, the model training configuration, and the evaluation metrics used to assess both building localization and wildfire vulnerability classification.

\subsection{Study Area}
\begin{figure}[ht]
    \centering
    \includegraphics[width=\textwidth]{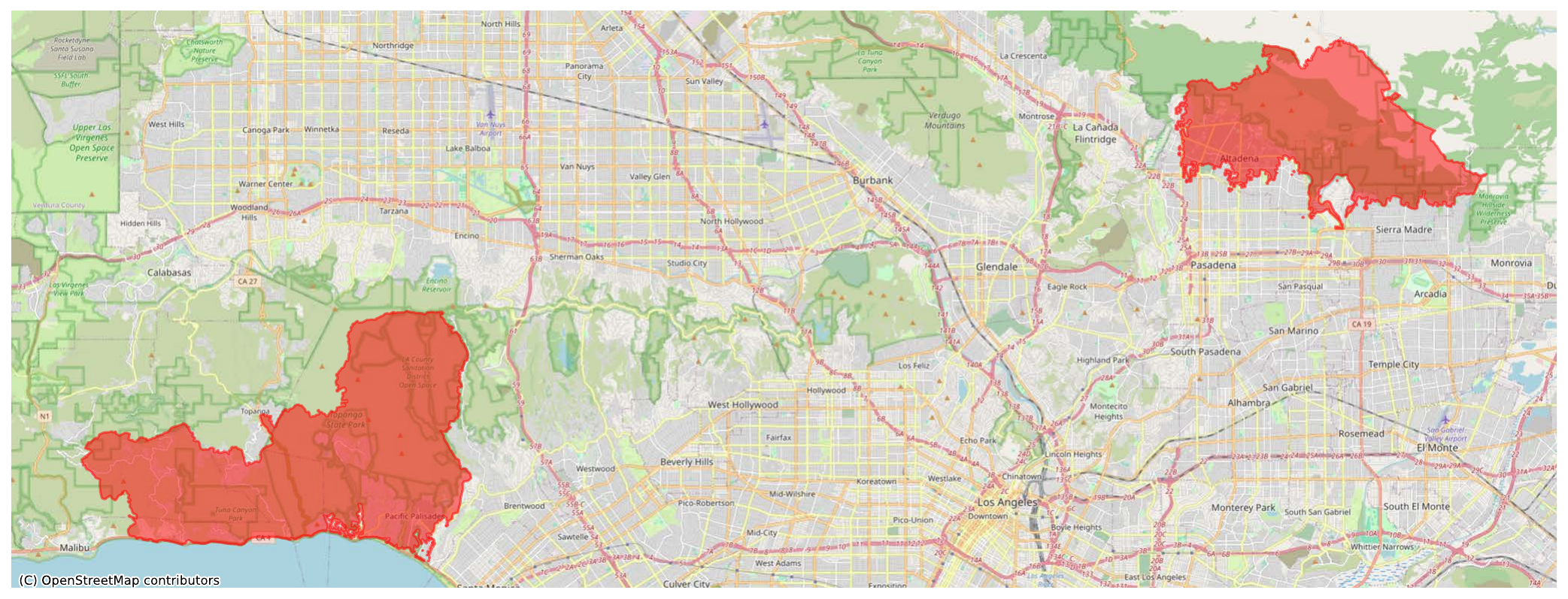}
    \caption{Study area: final perimeters of the 2025 Eaton and Palisades wildfires in the Los Angeles region, showing the wildland-urban interface neighborhoods used for structure-level vulnerability evaluation. Perimeter data are sourced from the City of Los Angeles GeoHub \citep{lafire_zone_2025}.}
    \label{fig:study_area}
\end{figure}

The validation of the proposed framework was conducted using case studies centered on the 2025 Eaton Fire and the Palisades Fire in Los Angeles, California, which both ignited on January 7, 2025, and burned until their containment on January 31, 2025. The study area is shown in Figure \ref{fig:study_area}. These highly destructive wildland-urban interface fires were characterized by rapid spread across complex, steep topographies, leading to extensive and varied degrees of structural damage across residential neighborhoods. The selection of these major WUI fire events offers a robust and challenging real-world scenario for evaluating the capabilities of advanced, structure-level wildfire vulnerability assessment methodologies. Critically, the availability of comprehensive ground truth data from post-event structural damage inspections for these specific regions, was a decisive factor in their selection for this research. The multimodal datasets subsequently employed for model training and evaluation are detailed below.

\subsection{Datasets}
\subsubsection{Satellite Imagery}
The primary overhead visual inputs consisted of WorldView-3 pansharpened optical imagery with an approximately 0.30-m ground sampling distance. To ensure comprehensive coverage of the affected regions prior to the disaster events, the imagery was acquired through a combination of the Maxar Open Data Program \citep{maxar_open_data_2026} and Apollo Mapping \citep{apollo_worldview3_2026}. All satellite images underwent orthomosaic processing to correct for topographic relief and camera tilt. For the Eaton Fire study area, pre-event imagery was captured on January 1, 2025. For the Palisades Fire study area, the pre-event imagery was acquired on October 20, 2024 and December 21, 2024. These pre-event acquisitions provided the foundational, macro-level landscape and structural context necessary for the assessments. In total, the study area was divided into 481 high-resolution $1024 \times 1024$ satellite image patches.

\subsubsection{Street-View Imagery}
To capture the critical facade-level vulnerabilities that are invisible from overhead perspectives, ground-level visual data were acquired from Google Street View \citep{google_street_view}. For each identified building footprint within the study areas, panoramic street-view images were systematically retrieved. The data acquisition was temporally constrained to fetch the most recent available image captured between the years 2020 and 2024 for each structure. This five-year pre-event window was strategically selected to balance temporal accuracy with maximum spatial coverage. Because street-level mapping vehicles do not traverse every residential WUI road on an annual basis, utilizing a multi-year window ensures that a ground-level perspective is successfully retrieved for the vast majority of structures in the dataset while remaining recent enough to accurately represent the structural materials, siding types, and immediate defensible space present at the time of the disaster. Alongside the visual data, critical metadata encompassing the camera's geographical coordinates and its heading direction in azimuth were recorded. This metadata serves as the foundational input for the geometric projection algorithm required to spatially anchor the ground-level attributes to the satellite basemap. This acquisition process resulted in a total of 16,577 localized panoramic street-view images.

\subsubsection{Ground Truth Data and Risk Derivation}
\label{sec:groung truth}
The baseline ground truth data for post-event structural impacts was obtained from the CAL FIRE Damage Inspection (DINS) Database \citep{calfire_dins_data_2026}. The raw DINS dataset categorizes structural damage into five ordinal levels based on the percentage of physical destruction. The initial raw distribution consisted of 17,209 structures with no damage (0\%), 1,847 affected (1–9\%), 467 with minor damage (10–25\%), 191 with major damage (26–50\%), and 17,264 destroyed (\textgreater50\%). 

However, utilizing absolute damage as the target variable for a vulnerability assessment presents a significant challenge. Absolute damage is heavily dominated by macroscopic fire spread dynamics, such as wind patterns or terrain-driven fire fronts. A destroyed building might simply have been in the direct path of an overwhelming firestorm rather than possessing inherently poor defensible space or vulnerable building materials, thereby obscuring its true, individual structural vulnerability. 

To isolate the inherent vulnerability of individual structures from these macroscopic fire behaviors, a relative risk scoring methodology was implemented. First, a spatially constrained agglomerative clustering approach, utilizing Ward's minimum variance criterion \citep{ward1963hierarchical} and implemented via the \texttt{spopt} Python library \citep{spopt2022}, was deployed to group the dataset into defined local neighborhoods based on geographic proximity and damage extent similarity. Within each neighborhood, the localized majority damage level was calculated to establish a baseline of the localized fire intensity. Next, a feature transformation was applied based on the premise that structures sustaining higher damage than their immediate neighbors possess a higher inherent risk, whereas structures that survive alongside destroyed neighbors demonstrate high structural resilience.

This relative risk score was calculated by determining the discrete deviation ($\Delta$) between an individual structure's ordinal damage class and its local neighborhood's majority damage class. To preserve the ordinal granularity of the original five-tier inspection data while ensuring robust statistical representation, these deviations were symmetrically capped and mapped back onto a newly defined, five-tier ordinal wildfire risk scale. Structures that sustained damage perfectly consistent with their local micro-environment ($\Delta = 0$) were designated as moderate risk (Class 2). Structures exhibiting higher comparative vulnerability were classified as high risk (Class 3) for a deviation of $\Delta = 1$, while any deviation of $\Delta \ge 2$ was grouped into extreme risk (Class 4). This upper cap prevents extreme outliers from fracturing the dataset into overly sparse, unlearnable categories. Conversely, structures demonstrating higher resilience than their neighbors were assigned to low risk (Class 1) for $\Delta = -1$, and minimal risk (Class 0) for all deviations of $\Delta \le -2$. This spatial aggregation and transformation process shifted the target variables from absolute damage toward true structural vulnerability. The final preprocessed dataset utilized for model training and evaluation yielded 8,224 minimal risk, 4,072 low risk, 6,807 moderate risk, 7,441 high risk, and 10,434 extreme risk structures. These refined labels serve as the ground truth for the framework's risk classification task.

\subsubsection{Building Footprints}
The delineation of building boundaries is essential for both the building localization task and the spatial aggregation of pixel-wise risk predictions. Georeferenced building footprint vector data were acquired from the County of Los Angeles Open Data portal \citep {lacounty_buildings_2026}. These vector footprints served two primary functions within the experimental framework. First, they were rasterized to generate the binary ground truth building masks utilized to train and evaluate the model's building localization function. Second, these geographic boundaries were employed during the inference and evaluation phases to spatially group the model's pixel-level risk classifications, thereby computing the building-level metric and generating the final actionable wildfire risk maps.

\subsection{Data Preparation and Augmentation}

A spatial pairing process was executed to associate the satellite image tiles with their corresponding discrete street-view text descriptions. First, the algorithm isolated only the textual data points that geographically intersected the bounding box of a given image tile. Then, the real-world geographic coordinates of these intersecting points were converted into 2D pixel coordinates relative to the image tensor. The extracted text features and their calculated pixel coordinates were then compiled into structured JSON files, establishing a many-to-one mapping between the localized ground-level attributes and each overhead basemap tile.

Following this pairing process, the dataset was partitioned into training (60\%), validation (20\%), and testing (20\%) subsets. To strictly prevent spatial data leakage, standard random sampling was intentionally avoided during dataset partitioning. Instead, a spatial splitting method was employed while maintaining a similar damage data distribution across subsets, ensuring rigorously isolated training, validation, and testing subsets. During the training phase, $512 \times 512$ pixel tiles were randomly cropped from the satellite image tiles and their corresponding ground truth masks. To enhance the model's generalization capabilities, these training inputs were subjected to spatial augmentations, including random rotations, horizontal flips, and vertical flips. Additionally, standard normalization procedures were applied to all input imagery prior to processing.

Because the proposed framework explicitly aligns these localized street-view text descriptions with specific geographic locations within the satellite basemap, any geometric transformation applied to the visual inputs must be synchronized with the point coordinates of the text descriptions. To maintain this spatial integrity, custom augmentation functions were implemented to recalculate the pixel coordinates of the street-view text descriptions alongside the image tensors.

For an image tile of height $h$ and width $w$, let the original pixel coordinate of a street-view text feature be defined as $(x, y)$. When an image is flipped horizontally, the horizontal coordinate is inverted such that the augmented coordinates become $(x', y') = (w - 1 - x, y)$. Similarly, when an image undergoes a vertical flip, the vertical coordinate is inverted to yield $(x', y') = (x, h - 1 - y)$. For counter-clockwise rotations applied in $k \times 90^\circ$ increments, where $k \in \{0, 1, 2, 3\}$, the coordinates transpose and invert based on the specific degree of rotation. A $90^\circ$ rotation transforms the point to $(x', y') = (y, w - 1 - x)$, a $180^\circ$ rotation maps it to $(x', y') = (w - 1 - x, h - 1 - y)$, and a $270^\circ$ rotation results in $(x', y') = (h - 1 - y, x)$.

Furthermore, during random cropping, if the original image requires zero-padding to meet the target $512 \times 512$ crop size, the text coordinates are first shifted by the calculated padding offsets $(W_{pad}, H_{pad})$ to derive their padded coordinates $(x_{pad}, y_{pad})$. The framework then filters the text features, retaining only the text features that fall within the boundaries of the randomly generated crop box originating at $(W_{start}, H_{start})$. For these retained points, the final localized coordinates are adjusted relative to the new crop window: $(x', y') = (x_{pad} - W_{start}, y_{pad} - H_{start})$.
This spatial tracking guarantees that the localized structural vulnerabilities identified in the street-view text are precisely anchored to their corresponding visual positions across all training epochs.

\subsection{Training Configuration}
The proposed vision-language framework was optimized using the AdamW optimizer. To preserve the robust, generalized feature representations learned during the DINOv3 foundation model's pretraining, the backbone network was completely frozen. Consequently, only the newly initialized parameters within the adapter layers and decoder heads were actively trained, utilizing an initial learning rate of $1 \times 10^{-4}$. A weight decay of $1 \times 10^{-2}$ was applied across all trainable parameters. These specific hyperparameter values were empirically determined through preliminary experiments to optimize model performance and stabilize the training trajectory.

For the composite loss function, the weights for the cross-entropy loss $\lambda_{ce}$ for both the building localization and wildfire vulnerability assessment tasks were set to $1.0$. The weight for the Lovasz-Softmax loss applied to the building localization task $\lambda_{lov}^{loc}$ was set to $0.5$. Recognizing the multi-class vulnerability assessment as the primary, more complex objective of the framework, the weight for its corresponding Lovasz-Softmax loss component $\lambda_{lov}^{clf}$ was assigned a higher value of $0.75$ to further heavily penalize IoU errors in the minority risk classes. 

All models were trained for a total of 100 epochs with an effective batch size of 16. Due to the high GPU memory demands associated with processing high-resolution visual inputs and large language feature embeddings, this was implemented using a true batch size of 2 complemented by 8 gradient accumulation steps. During the testing and inference phase, the model processed the input data at its original, uncropped resolution of $1024 \times 1024$ pixels to generate the final structural vulnerability predictions.

\subsection{Evaluation Metrics}
Model performance is evaluated using F1 scores, which provide a balanced assessment of precision and recall. Precision measures the accuracy of the positive predictions, while recall measures the model's ability to identify all actual positive instances. They are defined as
$$Precision = \frac{TP}{TP + FP}, \quad Recall = \frac{TP}{TP + FN}$$
where $TP$, $FP$, and $FN$ represent the number of true positives, false positives, and false negatives, respectively. The standard F1 score is then defined as the harmonic mean of these two metrics:
$$F1 = 2 \cdot \frac{Precision \cdot Recall}{Precision + Recall}$$
The building footprint extraction is evaluated using a binary pixel-level F1 score ($F1_{loc}$). For the wildfire risk assessment, the F1 score is calculated independently for each risk class. To provide a single, robust classification metric, these individual scores are aggregated using a harmonic mean F1 score ($F1_{clf}$). By using a harmonic mean rather than an arithmetic mean, the metric penalizes models that perform poorly on minority classes. This aggregation is defined as
$$F1_{clf} = \frac{|C|}{\sum_{c=1}^{|C|} \frac{1}{F1_c}}$$
where $|C|$ is the total number of evaluated risk classes, and $F1_c$ represents the individual F1 score for class $c$. If any class-specific F1 score is zero, the harmonic mean is defined as zero to avoid division by zero and to reflect complete failure on at least one class.

Finally, to automate best-model selection during the validation phase, a comprehensive overall validation score ($S_{val}$) is computed using a weighted sum that prioritizes risk classification over footprint localization:
$$S_{val} = 0.3 \cdot F1_{loc} + 0.7 \cdot F1_{clf}$$
During training, the model that achieves the highest $S_{val}$ on the validation dataset is preserved as the optimal model.

Finally, to translate these validated pixel-wise predictions into actionable, structure-level insights for operational evaluation, building-level F1 scores are subsequently computed by aggregating the pixel predictions within each building footprint using a majority vote. Because wildfire risk severity is an ordinal variable, standard F1 scores can be overly rigid by treating all misclassifications as equally incorrect. To account for this ordinal hierarchy at the structure level, the evaluation introduces an adjacent F1 score, which credits building-level predictions that fall within exactly one class level ($+1$ or $-1$) of the true ground truth label. Adjacent Precision ($Precision_{adj}$) and adjacent recall ($Recall_{adj}$) for a specific class $c$ are mathematically formulated using an indicator function $I$:
$$Precision_{adj} = \frac{\sum I(\hat{y}_i = c \land y_i \in \{c-1, c, c+1\})}{\sum I(\hat{y}_i = c)}$$
$$Recall_{adj} = \frac{\sum I(y_i = c \land \hat{y}_i \in \{c-1, c, c+1\})}{\sum I(y_i = c)}$$
where $y_i$ is the ground truth risk class and $\hat{y}_i$ is the predicted risk class for building $i$.
Furthermore, in disaster risk assessment, slightly overestimating vulnerability is generally preferable to underestimating it. To reflect this operational priority, an upper adjacent F1 score is utilized. This metric specifically credits predictions that either match the true class exactly or overestimate the risk by exactly one class level:
$$Precision_{u-adj} = \frac{\sum I(\hat{y}_i = c \land y_i \in \{c-1, c\})}{\sum I(\hat{y}_i = c)}$$
$$Recall_{u-adj} = \frac{\sum I(y_i = c \land \hat{y}_i \in \{c, c+1\})}{\sum I(y_i = c)}$$
The respective adjacent and upper adjacent F1 scores are then calculated using the standard harmonic mean of these modified precision and recall values.

\section{Results}
\label{sec:results}
The experimental evaluation examines the proposed framework from four complementary perspectives. First, three geolocation embedding strategies are compared to determine how localized street-view semantics can be most effectively anchored within overhead satellite features. Second, the predictive performance of the selected configuration is evaluated at both pixel and building levels. Third, ablation experiments quantify the contributions of satellite imagery, street-view semantic information, and explicit geolocation encoding. Finally, a controlled label-reconstruction experiment isolates the model’s spatial-alignment capability from limitations in the semantic information extracted from real-world street-view imagery.

\subsection{Comparison of Geolocation Embeddings}
To determine the optimal strategy for encoding the geographic coordinates of the street-view text descriptions into the multimodal architecture, a comparative analysis was conducted evaluating three distinct geolocation embedding techniques: Rotary Position Embedding, Random Fourier Features, and Spatial Grid Mapping. The quantitative results of this comparison are summarized in Table \ref{tab:geolocation_embeddings}. The RoPE strategy emerged as the significantly superior approach, achieving the highest overall harmonic mean F1 score of 0.2476 for the building wildfire risk classification task and the best building localization F1 score of 0.6735. Crucially, RoPE demonstrated the most robust capability in identifying critical structural vulnerabilities, dominating the moderate, high, and extreme risk categories. 

In contrast, while the Random Fourier Features method matched RoPE's computational efficiency at 100 million parameters and provided highly competitive building localization performance with F1 score of 0.6699, it generally yielded lower predictive accuracy across the more severe risk categories, resulting in a reduced harmonic mean of 0.2198. The Spatial Grid Mapping method proved to be suboptimal for this architecture. Despite significantly inflating the model's complexity to 349 million trainable parameters, it suffered a catastrophic failure in learning the moderate risk class, yielding an F1 score of zero which subsequently collapsed the overall harmonic mean to zero. Furthermore, its building localization performance degraded severely to 0.2284. These findings clearly validate the selection of RoPE as the primary geolocation embedding mechanism.

\begin{table}[htbp]
\centering
\caption{Performance comparison of geolocation embedding strategies. Best results are bolded, second-best are underlined.}
\label{tab:geolocation_embeddings}
\resizebox{\textwidth}{!}{%
\begin{tabular}{l c c c c c c c c}
\toprule
\multirow{2}{*}{Geolocation Embedding} & \multirow{2}{*}{\begin{tabular}[c]{@{}c@{}}Trainable \\ Param \# \\(M)\end{tabular}} & \multirow{2}{*}{\begin{tabular}[c]{@{}c@{}}Building\\Localization\\ F1 Score\end{tabular}} & \multicolumn{6}{c}{Building Wildfire Risk Classification F1 Score} \\ \cline{4-9} 
 &  &  & Minimal & Low & Moderate & High & Extreme & \begin{tabular}[c]{@{}c@{}}Harmonic\\ Mean\end{tabular} \\ \midrule
Rotary Position Embedding & 100 & \textbf{0.6735} & 0.2330 & 0.2147 & \textbf{0.1747} & \textbf{0.6842} & \textbf{0.2462} & \textbf{0.2476} \\
Random Fourier Features & 100 & \underline{0.6699} & \underline{0.3588} & \underline{0.2296} & \underline{0.1048} & \underline{0.5320} & \underline{0.2385} & \underline{0.2198} \\
Spatial Grid Mapping & 349 & 0.2284 & \textbf{0.3916} & \textbf{0.4012} & 0.0000 & 0.4027 & 0.2267 & 0.0000 \\ \bottomrule
\end{tabular}%
}
\end{table}

\subsection{Structure-Level Risk Prediction Performance}
\begin{figure}[!ht]
    \centering
    \includegraphics[width=0.7\textwidth]{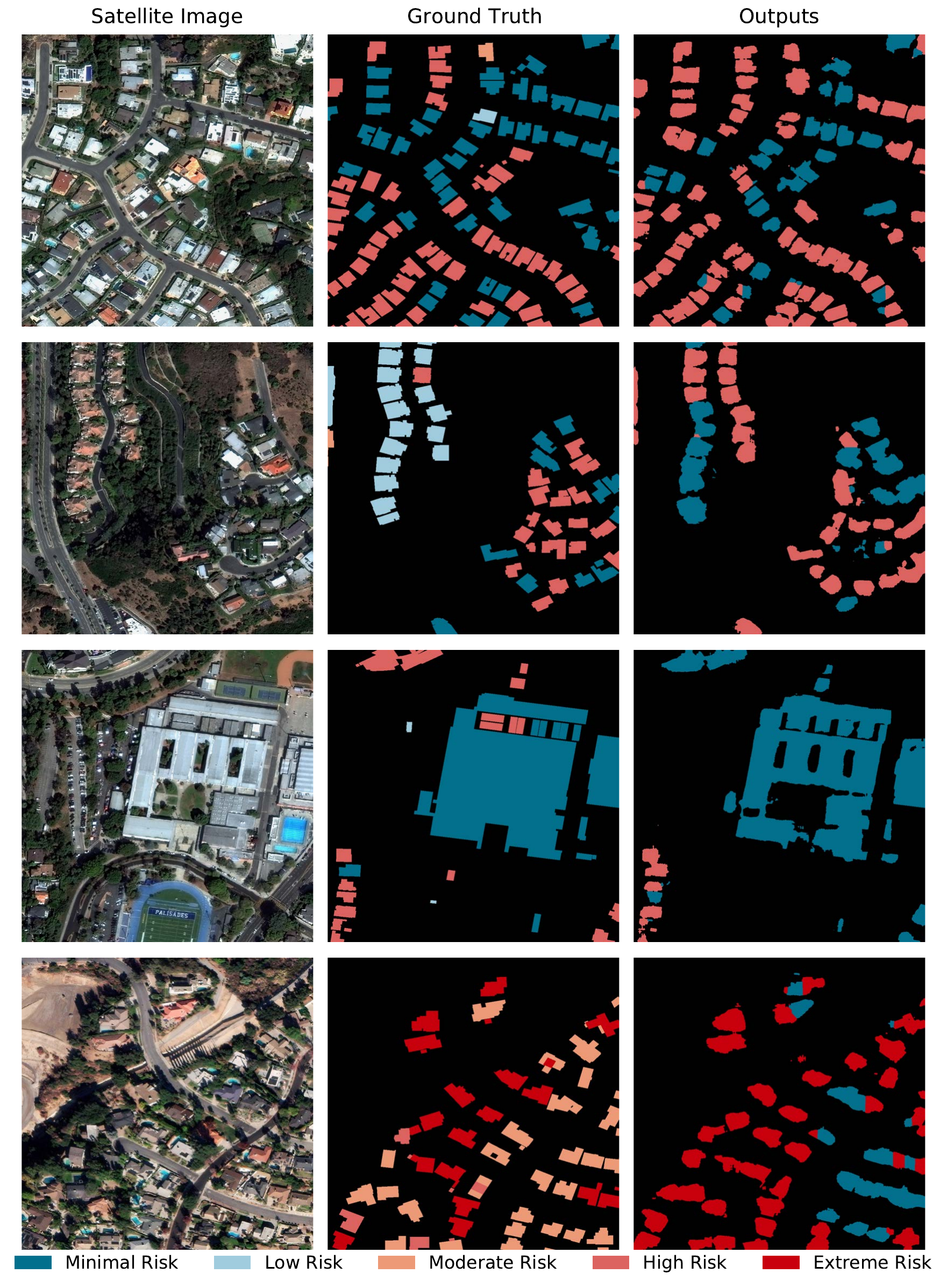}
    \caption{Pixel-level wildfire vulnerability predictions from the proposed RoPE-based multimodal framework (right) compared against the ground-truth building-level risk labels (left) for representative tiles in the Eaton and Palisades study areas. Colors denote the five ordinal risk classes from minimal to extreme; the comparison illustrates the model's capacity to capture broad spatial patterns of structural vulnerability while highlighting residual confusion between adjacent low and moderate tiers.}
    \label{fig:wildfire_risk_visualization}
\end{figure}

To qualitatively assess the framework's predictive capabilities, Figure \ref{fig:wildfire_risk_visualization} visualizes the pixel-level risk predictions generated by the optimal model utilizing RoPE geolocation embeddings, alongside their corresponding ground-truth localized building risk maps. As observed in the visual comparisons, the model effectively captures the broad spatial distribution of structural vulnerabilities, demonstrating a particularly strong capability in delineating minimal, high, and extreme risk clusters. However, structures within the low and moderate risk tiers are more susceptible to misclassification, a limitation likely attributable to the lack of highly distinguishable visual and textual characteristics differentiating these intermediate vulnerability levels.

For a rigorous quantitative evaluation, the model's performance was assessed across both the raw pixel outputs and building-level aggregations. Pixel-level F1 scores evaluate the direct output of the multimodal decoder. However, because wildfire vulnerability is fundamentally a structure-level attribute, the pixel predictions were subsequently aggregated within the boundaries of the building footprints using majority voting to calculate building-level F1 scores. Recognizing the inherent ordinal nature of the wildfire risk scale, we also report adjacent F1 and upper adjacent F1 scores. The comprehensive results across these metrics are detailed in Table \ref{tab:f1_results}.

\begin{table}[htbp]
\centering
\caption{Quantitative evaluation of building wildfire risk classification across pixel-level and building-level aggregated metrics. Best results are bolded, second-best are underlined.}
\label{tab:f1_results}
\resizebox{\textwidth}{!}{%
\begin{tabular}{lcccccc}
\toprule
Metric & Minimal & Low & Moderate & High & Extreme & Harmonic Mean \\ 
\midrule
Pixel-level F1 & \textbf{0.2330} & \underline{0.2147} & 0.1747 & 0.6842 & 0.2462 & 0.2476 \\
Building-level F1 & 0.1561 & 0.1993 & 0.2241 & 0.7481 & 0.2173 & 0.2594 \\
Building-level Adjacent F1 & \underline{0.2177} & \textbf{0.5058} & \textbf{0.2717} & \textbf{0.7705} & \textbf{0.2937} & \textbf{0.3861} \\
Building-level Upper Adjacent F1 & 0.1645 & 0.2146 & \underline{0.2628} & \underline{0.7670} & \underline{0.2884} & \underline{0.3022} \\ 
\bottomrule
\end{tabular}%
}
\end{table}

The results in Table \ref{tab:f1_results} reveal critical insights into the framework's error distribution and predictive bias. By analyzing the differences between the strict, adjacent, and upper adjacent metrics, distinct behavioral tendencies emerge across the risk scale. Crucially, the framework exhibits a conservative, fail-safe bias in the higher vulnerability tiers. For both the moderate and extreme classes, the upper adjacent metric accounts for the vast majority of the improvement observed in the full adjacent score. For the moderate class, the score improves from a strict F1 of 0.2241 to an upper adjacent F1 of 0.2628, capturing almost all of the full adjacent score's total increase to 0.2717. Similarly, the extreme class jumps from a strict F1 of 0.2173 to an upper adjacent score of 0.2884, which closely mirrors its maximum full adjacent score of 0.2937. This signifies that when the model struggles to perfectly classify a structure, it predominantly errs on the side of caution by overestimating the risk. Specifically, because the extreme class is the highest possible category, the notable improvement in its upper adjacent score mathematically confirms that the model frequently flags high-risk ground-truth structures as extreme. Similarly, moderate structures are more likely to be conservatively elevated to the high-risk category than downgraded. Despite this conservative bleed-over into the extreme class, the high-risk category demonstrates consistently robust performance across all three metrics, maintaining a strict F1 of 0.7481, an upper adjacent F1 of 0.7670, and a full adjacent F1 of 0.7705. In the context of disaster management, confidently isolating high-risk structures while being conservative via over-prediction in ambiguous intermediate cases is a highly desirable characteristic. Conversely, an under-prediction bias is evident at the lower end of the scale. For the low-risk category, the F1 score drastically improves from 0.1993 to 0.5058 under the full adjacent constraint, yet the upper adjacent F1 only reaches 0.2146. This reveals that the model frequently misses the subtle visual and textual cues distinguishing minimal from low risk, causing it to systematically misclassify low-risk structures as minimal.

\subsection{Ablation Studies}
To rigorously evaluate the contribution of each architectural component within the proposed framework, comprehensive ablation studies were conducted. Specifically, the experiments aimed to isolate the individual impacts of the input modalities as well as the efficacy of the geometric spatial alignment provided by the Rotary Position Embedding mechanism. The quantitative results of these configurations are detailed in Table \ref{tab:ablation_studies}.

The baseline unimodal models demonstrated expected limitations when operating in isolation. The model utilizing only satellite imagery achieved a competitive building localization F1 score of 0.6527, leveraging the clear geometric boundaries visible from overhead perspectives. However, it struggled significantly with the building wildfire risk classification task, yielding a low harmonic mean F1 of 0.1450. Conversely, the unimodal model relying solely on unanchored street-view text descriptions completely failed at localization and classification, with F1 scores of 0.0884 and 0.0764, respectively, underscoring that semantic descriptions without spatial context are insufficient. When RoPE was introduced to the text-only model, classification performance rebounded dramatically to a harmonic mean of 0.1964, proving that spatially structuring the text is critical without visual data.

The advantages of the multimodal approach become evident when both data sources are fused. Simply integrating the satellite imagery and text descriptions without geographic embeddings improved the overall risk classification harmonic mean to 0.2115, outperforming both unimodal baselines. However, the full potential of the framework was only realized when the multimodal inputs were coupled with the RoPE geolocation embeddings. This finalized architecture achieved the highest overall performance across both tasks, achieving the highest harmonic mean F1 score of 0.2476 and boosting building localization to an F1 score of 0.6735. These results confirm the core hypothesis of this research that overhead imagery provides the essential macro-level geometric context, but spatially anchoring ground-level semantic vulnerabilities via precise geometric projection is necessary to accurately assess structure-level wildfire risk. 

\begin{table}[htbp]
\centering
\caption{Ablation study results on input modalities and geolocation embeddings. Best results are bolded, second-best are underlined.}
\label{tab:ablation_studies}
\resizebox{\textwidth}{!}{%
\begin{tabular}{l c c c c c c c c}
\toprule
\multirow{2}{*}{Model} & \multirow{2}{*}{\begin{tabular}[c]{@{}c@{}}Trainable \\ Param \# \\(M)\end{tabular}} & \multirow{2}{*}{\begin{tabular}[c]{@{}c@{}}Building\\Localization\\ F1 Score\end{tabular}} & \multicolumn{6}{c}{Building Wildfire Risk Classification F1 Score} \\ \cline{4-9} 
 &  &  & Minimal & Low & Moderate & High & Extreme & \begin{tabular}[c]{@{}c@{}}Harmonic\\ Mean\end{tabular} \\ \midrule
 
Satellite Imagery & 58 & \underline{0.6527} & 0.0595 & \underline{0.4181} & 0.1362 & \textbf{0.6976} & 0.1529 & 0.1450 \\[1.5ex]

Street-view Text Descriptions & \multirow{2}{*}{100} & \multirow{2}{*}{0.0884} & \multirow{2}{*}{0.0445} & \multirow{2}{*}{0.0763} & \multirow{2}{*}{0.0517} & \multirow{2}{*}{0.5168} & \multirow{2}{*}{0.1159} & \multirow{2}{*}{0.0764} \\
(w/o Geolocation Embedding) & & & & & & & & \\[1.5ex]

Street-view Text Descriptions & \multirow{2}{*}{100} & \multirow{2}{*}{0.3296} & \multirow{2}{*}{\underline{0.2441}} & \multirow{2}{*}{0.2361} & \multirow{2}{*}{\textbf{0.2267}} & \multirow{2}{*}{0.4376} & \multirow{2}{*}{0.0958} & \multirow{2}{*}{0.1964} \\
(Geolocation Embedding: RoPE) & & & & & & & & \\[1.5ex]

Satellite Imagery + Street-view Text Descriptions & \multirow{2}{*}{100} & \multirow{2}{*}{0.6453} & \multirow{2}{*}{\textbf{0.2765}} & \multirow{2}{*}{\textbf{0.4292}} & \multirow{2}{*}{0.0849} & \multirow{2}{*}{0.6399} & \multirow{2}{*}{\underline{0.2296}} & \multirow{2}{*}{\underline{0.2115}} \\
(w/o Geolocation Embedding) & & & & & & & & \\[1.5ex]

Satellite Imagery + Street-view Text Descriptions & \multirow{2}{*}{100} & \multirow{2}{*}{\textbf{0.6735}} & \multirow{2}{*}{0.2330} & \multirow{2}{*}{0.2147} & \multirow{2}{*}{\underline{0.1747}} & \multirow{2}{*}{\underline{0.6842}} & \multirow{2}{*}{\textbf{0.2462}} & \multirow{2}{*}{\textbf{0.2476}} \\
(Geolocation Embedding: RoPE) & & & & & & & & \\ \bottomrule

\end{tabular}%
}
\end{table}

\subsection{Label Reconstruction Diagnostic Experiment}

While the ablation studies demonstrated that the Rotary Position Embedding improved the framework's performance, the magnitude of the improvement provided by the geolocation embeddings on the primary task was less significant than anticipated. To investigate whether this limitation stemmed from the architectural mechanism of the spatial embeddings or from the inherent nature of the street-view text descriptions themselves, a label reconstruction diagnostic experiment was conducted. 

In this controlled experiment, the street-view text descriptions were replaced with explicit ground-truth risk level labels passed as the text input. The objective was to determine if the model could successfully reconstruct the building-level vulnerability maps when given perfect textual information, thereby isolating the model's capacity to spatially anchor text to visual features. The quantitative results are presented in Table \ref{tab:label_reconstruction}.

In contrast to the primary task, explicit geolocation encoding produced substantially larger gains in the label-reconstruction experiment. Without geolocation embeddings, the model achieved a harmonic mean F1 of 0.3458; this increased to 0.4466 with RoPE and 0.5329 with RFF. These results demonstrate that the fusion architecture can effectively exploit explicit spatial anchors when highly informative textual inputs are available. The smaller gains observed with real-world VLM-derived descriptions therefore likely reflect limitations in the semantic information extracted from street-view imagery rather than an inability of the model to utilize geolocation information.

\begin{table}[htbp]
\centering
\caption{Results of the label reconstruction diagnostic experiment utilizing explicit risk levels as text input. Best results are bolded, second-best are underlined.}
\label{tab:label_reconstruction}
\resizebox{\textwidth}{!}{%
\begin{tabular}{l c c c c c c c}
\toprule
\multirow{2}{*}{Model} & \multirow{2}{*}{\begin{tabular}[c]{@{}c@{}}Building\\Localization\\ F1 Score\end{tabular}} & \multicolumn{6}{c}{Building Wildfire Risk Classification F1 Score} \\ \cline{3-8} 
 & & Minimal & Low & Moderate & High & Extreme & \begin{tabular}[c]{@{}c@{}}Harmonic\\ Mean\end{tabular} \\ \midrule
 
Satellite Imagery + Risk Labels as Text & \multirow{2}{*}{\textbf{0.6806}} & \multirow{2}{*}{0.4748} & \multirow{2}{*}{\underline{0.3165}} & \multirow{2}{*}{\textbf{0.7516}} & \multirow{2}{*}{0.7878} & \multirow{2}{*}{0.1516} & \multirow{2}{*}{0.3458} \\
(w/o Geolocation Embedding) & & & & & & & \\[1.5ex]

Satellite Imagery + Risk Labels as Text & \multirow{2}{*}{\underline{0.6618}} & \multirow{2}{*}{\textbf{0.6584}} & \multirow{2}{*}{0.1976} & \multirow{2}{*}{\underline{0.4360}} & \multirow{2}{*}{\underline{0.9187}} & \multirow{2}{*}{\textbf{0.8082}} & \multirow{2}{*}{\underline{0.4466}} \\
(Geolocation Embedding: RoPE) & & & & & & & \\[1.5ex]

Satellite Imagery + Risk Labels as Text & \multirow{2}{*}{0.6420} & \multirow{2}{*}{\underline{0.6092}} & \multirow{2}{*}{\textbf{0.3309}} & \multirow{2}{*}{0.4261} & \multirow{2}{*}{\textbf{0.9267}} & \multirow{2}{*}{\underline{0.7727}} & \multirow{2}{*}{\textbf{0.5329}} \\
(Geolocation Embedding: RFF) & & & & & & & \\ \bottomrule
\end{tabular}%
}
\end{table}

\section{Discussion}
\label{sec:discussion}
The most innovative aspect of this study is that it shows how structure-level wildfire vulnerability can be inferred by spatially aligning semantically compressed ground evidence with continuous overhead imagery, rather than by forcing direct pixel-to-pixel fusion across viewpoints. Empirically, the results validate this design choice: the full multimodal model with Rotary Positional Encoding achieved the best localization score ($F1 = 0.6735$) and the strongest five-class risk classification performance (harmonic mean $F1 = 0.2476$), outperforming both the unimodal baselines and the unanchored multimodal configuration. The remainder of this section interprets why this geolocation-aware alignment prevailed over alternative methods, examines its operational significance for wildland--urban interface risk management, and delineates the boundaries of the current framework.

The comparative experiments, specifically the contrast between the real-world dataset (Table \ref{tab:geolocation_embeddings}) and the label reconstruction diagnostic (Table \ref{tab:label_reconstruction}), highlight the nuanced role of spatial embeddings in multimodal reasoning. While Random Fourier Features excelled in the diagnostic task, Rotary Position Embedding proved superior for the primary real-world application. This divergence stems directly from their underlying mathematical mechanisms. RFF is an additive method that injects high-frequency, absolute coordinate anchors directly into the feature vectors. While this acts as a highly effective spatial hash for memorizing locations when the input is a perfectly certain ground-truth label, adding this high-frequency positional wave data to real-world, noisy natural language vectors disrupts and washes out their subtle semantic meanings. Conversely, RoPE employs a multiplicative approach, rotating feature vectors in complex space. This preserves the delicate semantic magnitude of the nuanced street-view descriptions while seamlessly injecting spatial awareness based on relative distances, making it uniquely suited for complex multimodal reasoning. Furthermore, the overall magnitude of improvement from geolocation embeddings was less pronounced in the real-world task compared to the diagnostic task. This discrepancy is primarily driven by the presence of implicit spatial baselines. VLM-derived street-view descriptions inherently contain rich visual semantics that the multimodal cross-attention mechanisms can implicitly align with the corresponding overhead visual features. Because the vision backbone naturally establishes this correspondence, explicit geolocation embeddings act merely as a fine-grained spatial refinement rather than building spatial awareness from scratch. Conversely, the diagnostic target labels contain no visual semantic information. Without these implicit visual cues to guide the attention routing, the diagnostic model relies entirely on the explicit geolocation embeddings to pin the text to the geographic grid, resulting in a massive performance spike when they are introduced. Finally, real-world performance is naturally constrained by the purity of the input signal; the street-view text is inherently noisy, occasionally ambiguous, capping the overall performance ceiling compared to the purely spatial memorization exercise utilizing 100\% certain ground-truth labels.

Despite the inherent noise within real-world street-view descriptions and the complexity of cross-modal alignment, the proposed multimodal fusion successfully achieves an operational solution highly suited for disaster management. Beyond predictive reliability, the text-based dimensionality reduction allows the framework to demonstrate exceptional scalability. The core multimodal architecture achieves a pure inference speed of 3.79 seconds per $1024 \times 1024$-pixels tile. Consequently, when executing the complete end-to-end operational pipeline sequentially with a batch size of 1, including data loading, model inference, and the generation of georeferenced risk maps, the framework assessed approximately 36,978 structures across the study area in just 37 minutes on a single NVIDIA RTX A6000. It is important to note that this unoptimized, sequential measurement is reported specifically to establish a baseline performance floor; in a true operational deployment, leveraging parallel processing and larger batch sizes would significantly reduce this execution time. By aggregating pixel predictions to structural boundaries, the model mitigates localized noise and demonstrates a valuable conservative bias. As discussed in the experimental results, the framework confidently isolates high-risk structures while tending to over-predict vulnerability in ambiguous, intermediate cases rather than under-predicting it. In the context of community resilience planning, this combination of rapid, large-scale processing and a fail-safe margin ensures that emergency managers and policymakers can efficiently triage resources across expansive WUI environments without overlooking highly susceptible homes.

This study should be understood as establishing the scope and feasibility of a new multimodal assessment paradigm rather than exhausting the problem of structure-level wildfire vulnerability. The first boundary is data availability: street-view coverage is spatially uneven, historical captures may be temporally misaligned with pre-event satellite imagery, and occlusions from trees, walls, or parked vehicles can hide the facade conditions the model seeks to assess. The second boundary is label purity: although the paper transforms post-fire damage into a relative vulnerability proxy, the target variable still inherits noise from fire behavior, suppression dynamics, and neighborhood effects that cannot be fully disentangled from pre-fire structure susceptibility. A third limitation is signal granularity, particularly for adjacent low and moderate classes where semantic cues are subtle and class distinctions may be harder to observe from available imagery. Future research should therefore incorporate richer pre-fire inspection datasets, extend the framework with complementary modalities such as parcel records or fuels data, and test transferability across different regions, building typologies, and fire regimes. Additional work on uncertainty quantification and temporal updating would also strengthen the framework's readiness for operational deployment.

\section{Concluding Remarks}
\label{sec:conclusion}
This study presents a novel framework for structure-level wildfire vulnerability assessment that overcomes the computational bottlenecks of community-scale multimodal fusion. Methodologically, it advances remote sensing analytics by (i) utilizing Vision-Language Models to compress high-resolution street-view imagery into semantically dense, lightweight text representations, (ii) leveraging a satellite-pretrained vision foundation model as the frozen backbone to extract macro-level spatial features from pre-event satellite imagery , and (iii) introducing a decoupled cross-attention decoder equipped with Rotary Position Embeddings to explicitly anchor numerous discrete and localized ground-level text features within a continuous overhead satellite basemap. This design replaces computationally prohibitive image-to-image fusion with semantically compressed, many-to-one alignment, reframing the multimodal problem around local coordinate grounding within a bounded satellite patch.

Empirically, the ablation and diagnostic experiments confirm that explicit spatial alignment is critical for bridging the gap between ground-level semantics and overhead imagery. The study reveals that RoPE, through its multiplicative coordinate transformations, effectively preserves delicate semantic details while injecting relative spatial awareness, outperforming additive methods like Random Fourier Features in real-world scenarios. Evaluating the model at the building level further reveals a highly desirable conservative bias in the upper vulnerability tiers, minimizing the dangerous under-prediction of severe risk.

From a practical perspective, the framework delivers a highly scalable operational pipeline for pre-fire screening of individual structures across expansive wildland-urban interface communities. By sidestepping the massive computational overhead of pixel-to-pixel cross-view fusion, the system successfully processed approximately 1,000 structures per minute on a single NVIDIA RTX A6000 GPU without parallel processing. Because the model can infer risk from pre-event satellite imagery and routinely available street-view captures, it supports mitigation prioritization before a fire occurs rather than only explaining damage after the event. Emergency managers and local governments could use the outputs to identify clusters of high-risk parcels, target defensible-space outreach, prioritize inspections, and stage resilience investments, while utilities, insurers, and risk modelers could leverage the structure-level outputs to supplement portfolio screening and community risk stratification where field inspections are infeasible at scale.

To translate the framework into practice, implementation should proceed as a staged workflow rather than a one-time modeling exercise. Agencies or partners would first assemble current building footprints, recent pre-event satellite imagery, and the most recent available street-view or equivalent ground-level imagery for the target jurisdiction; next, the Vision-Language Model extraction and coordinate projection modules would generate spatially anchored semantic descriptions for each structure-facing capture. The multimodal model could then produce tile-level predictions that are aggregated to building footprints and ranked into actionable intervention tiers for inspection, mitigation, or portfolio review. Operational uptake would require coordination among fire agencies, local governments, utilities, insurers, and community resilience offices, with clear protocols for how model outputs trigger field verification, homeowner outreach, or mitigation subsidies. The most likely barriers are incomplete ground imagery, uneven data refresh cycles, and institutional hesitation around black-box predictions; these can be mitigated through periodic data updates, fallback use of alternate ground imagery sources, and targeted review of extreme-risk or uncertain cases. Framed this way, the model becomes part of a decision-support pipeline: screen broadly, validate strategically, intervene selectively, and update continuously as new imagery becomes available.

\section*{Author Contribution}
Conceptualization: Dr. Yu-Hsuan Ho, Yiming Xiao, Dr. Hao-En Lai, Dr. Ali Mostafavi; Methodology: Dr. Yu-Hsuan Ho, Yiming Xiao, Dr. Hao-En Lai;  Investigation: Dr. Yu-Hsuan Ho; Supervision: Dr. Ali Mostafavi, Dr. Samuel D. Brody; Writing - Original Draft: Dr. Yu-Hsuan Ho; Writing - Review \& Editing: Yiming Xiao, Dr. Hao-En Lai, Dr. Ali Mostafavi, Dr. Samuel D. Brody.

\section*{Data Availability}
The baseline post-event structural damage ground truth data supporting the findings of this study are publicly available through the California Department of Forestry and Fire Protection (CAL FIRE) Damage Inspection (DINS) Database, accessible at \url{https://data.ca.gov/dataset/cal-fire-damage-inspection-dins-data}.

\section*{Code Availability}
The source code of the proposed model in this study can be found at \url{https://github.com/violayhho/Wildfire-SSR}.

\bibliography{ref}

\newpage
\section*{Supplementary Information}
\appendix

\setcounter{section}{0}
\renewcommand{\thesection}{S\arabic{section}}
\renewcommand{\thesubsection}{\thesection.\arabic{subsection}}

\setcounter{table}{0}
\renewcommand{\thetable}{S\arabic{table}}

\setcounter{figure}{0}
\renewcommand{\thefigure}{S\arabic{figure}}

\section{Street-View Semantic Feature Extraction Workflow}
\label{app:semantic_extraction}

This appendix details the pipeline utilized to convert raw ground-level street-view imagery into refined text descriptions. The workflow systematically progresses from the initial VLM prompting to data filtration and final natural language generation.

\subsection{Prompts for Vision-Language Models}
\label{app:vlm_prompt}
To consistently extract fine-grained building and environmental attributes, the GPT-4o Vision-Language Model was tasked with analyzing the panoramic street-view images. To prevent hallucination and ensure standardized formatting, the model was constrained to output its findings strictly as a JSON object based on a predefined schema. The exact prompt and schema provided to the VLM are as follows:

\begin{promptbox}
Analyze the wildfire-relevant structural and environmental attributes visible in the street-view image. For each attribute, select exactly one allowed value. Only values outside parentheses may appear in the output. Text inside parentheses is explanation only and must not be returned. If an attribute is not visible, occluded, or cannot be determined reliably, return "unsure". Use "other" only when the attribute is visible but does not match any listed category. Do not infer hidden or non-visible characteristics. Return only the JSON object in the following format :
{
  "roof material": "tile (clay/concrete); metal; asphalt shingle; wood shake; other; unsure",
  "roof condition": "clean; visible debris; damaged; unsure",
  "roof angle": "flat; 60; 90; 135; unsure",
  "debris in gutter": "true; false; unsure",

  "siding material": "stucco/cement; brick/stone; fiber cement; wood; vinyl; metal; other; unsure",
  "siding-to-ground clearance present": "true (foundation exposed); false (siding contacts ground/mulch); unsure",
  "building maintenance condition": "good; moderate; poor (peeling paint/rot); unsure",
  "structure type": "mobile home; wood; stone; rc; steel; other; unsure",

  "window type": "single-pane; multi-pane; shutters/boards installed; other; unsure",
  "vent type": "unscreened/open; standard screen; fine mesh screen; engineered baffle vent; other; unsure",
  "eave construction": "enclosed; open; unsure",

  "deck surface material": "wood; composite; non-combustible (concrete/stone); other; unsure",
  "under-deck condition": "enclosed; open and clean; open with debris/storage; unsure",
  "fence-to-house connection": "combustible fence attached; non-combustible fence attached; fence detached; no fence; unsure",
  "foundation type": "enclosed; open; unsure",

  "immediate zone ground cover": "non-combustible; bare soil; mowed lawn; combustible mulch; overgrown weeds/grass; other; unsure",
  "vegetation-structure contact": "direct contact; overhanging canopy; no contact; unsure",
  "vertical/ladder fuel continuity": "true; false; unsure",
  "overall vegetation condition": "green and well-maintained; dead/dry; dense and overgrown; other; unsure",

  "combustible outbuilding present": "true; false; unsure",
  "firewood pile/fuel tank adjacent to structure": "true; false; unsure",

  "terrain slope": "flat/gentle; moderate; steep; unsure",
  "adjacent parcel condition": "well-maintained; overgrown/unmanaged; unsure",
  "power lines": "vegetation near power lines; no vegetation near power lines; no power lines; unsure",
  "road access": "paved and wide (>=2 lanes); paved but narrow; unpaved; obstructed/overgrown; unsure"
}
\end{promptbox}

Following the execution of this prompt, the VLM generates a highly structured dictionary of the observed structural and landscape vulnerabilities. Section \ref{app:example_outputs} illustrates an application of this extraction step.

\subsection{Example Outputs}
\label{app:example_outputs}
Street-View Image:
\begin{figure}[ht]
    \centering
    \includegraphics[width=0.7\textwidth]{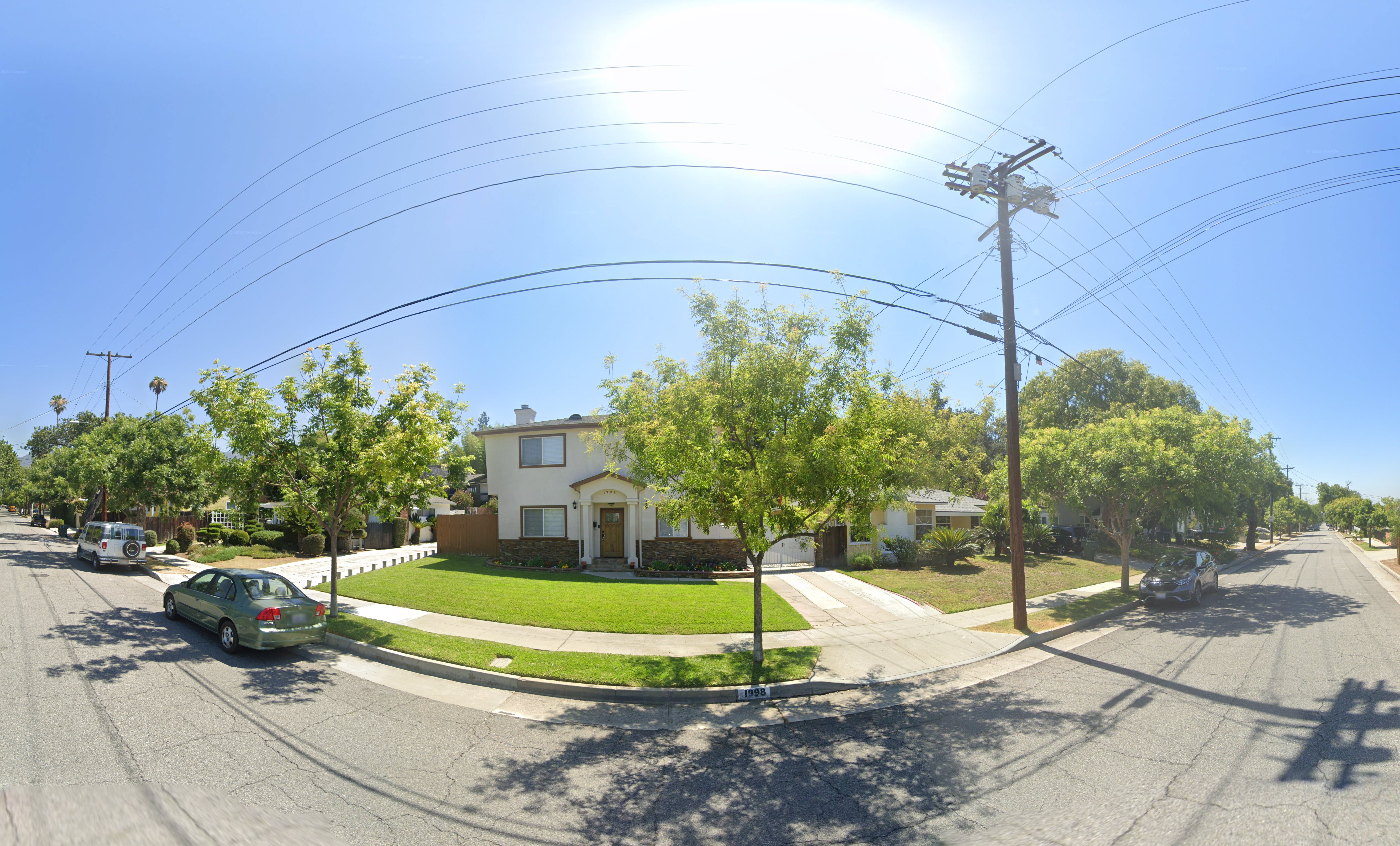}
    \caption{Example street-view image used for VLM-based semantic feature extraction.}
\end{figure}

VLM Output:
\begin{promptbox}
{
        "roof material": "tile",
        "roof condition": "clean",
        "roof angle": "135",
        "debris in gutter": "false",
        "siding material": "stucco/cement",
        "siding-to-ground clearance present": "false",
        "building maintenance condition": "good",
        "structure type": "wood",
        "window type": "multi-pane",
        "vent type": "standard screen",
        "eave construction": "enclosed",
        "deck surface material": "non-combustible",
        "under-deck condition": "enclosed",
        "fence-to-house connection": "combustible fence attached",
        "foundation type": "enclosed",
        "immediate zone ground cover": "mowed lawn",
        "vegetation-structure contact": "no contact",
        "vertical/ladder fuel continuity": "false",
        "overall vegetation condition": "green and well-maintained",
        "combustible outbuilding present": "false",
        "firewood pile/fuel tank adjacent to structure": "false",
        "terrain slope": "flat/gentle",
        "adjacent parcel condition": "well-maintained",
        "power lines": "vegetation near power lines",
        "road access": "paved and wide"
    }
\end{promptbox}

While the initial prompt instructs the model to evaluate 25 distinct characteristics, passing all variables directly into the text encoder introduces unnecessary semantic noise. To optimize the language embeddings for the downstream fusion decoder, the framework filters this raw output down to its most predictive components.

\newpage
\subsection{Wildfire Risk Feature Dictionary and Selection}
\label{app:feature_dictionary}

As discussed in Section \ref{sec: sv_extraction}, a recursive feature elimination strategy utilizing XGBoost was applied to isolate the variables that most significantly impact building survival. Table \ref{tab:feature_dictionary} details the complete dictionary of 25 features evaluated by the VLM. The variables marked with a ``Yes'' in the final column represent the 16 optimally selected contributing features that were retained for the final multimodal integration, successfully removing 9 non-contributing or overly noisy attributes.

\begin{table}[htbp]
    \centering
    \renewcommand{\arraystretch}{1.6} 
    \caption{Wildfire Risk Feature Table}
    \label{tab:feature_dictionary}
    \vspace{0.2cm}
    \begin{tabular}{llllc}
        \hline
        \textbf{ID} & \textbf{Feature Name} & \textbf{Primary Category} & \textbf{Sub-Category} & \textbf{\begin{tabular}[c]{@{}c@{}}Contributing\\Feature?\end{tabular}} \\
        \hline
        B-01 & Roof Material & Building Shell & Roof Assembly & Yes \\
        B-02 & Roof Condition & Building Shell & Roof Assembly & \\
        B-03 & Roof Angle & Building Shell & Roof Assembly & Yes \\
        B-04 & Debris in Gutter & Building Shell & Roof Assembly & \\
        B-05 & Siding Material & Building Shell & Vertical Surfaces & Yes \\
        B-06 & Siding-to-Ground Clearance Present & Building Shell & Vertical Surfaces & Yes \\
        B-07 & Building Maintenance Condition & Building Shell & Vertical Surfaces & \\
        B-08 & Structure Type & Building Shell & Structure & Yes \\
        B-09 & Window Type & Building Shell & Windows \& Openings & \\
        B-10 & Vent Type & Building Shell & Windows \& Openings & \\
        B-11 & Eave Construction & Building Shell & Eaves \& Overhangs & \\
        B-12 & Deck Surface Material & Attached Structures & Decks \& Porches & Yes \\
        B-13 & Under-Deck Condition & Attached Structures & Decks \& Porches & Yes \\
        B-14 & Fence-to-House Connection & Attached Structures & Fences & Yes \\
        B-15 & Foundation Type & Attached Structures & Foundation & \\
        S-01 & Immediate Zone Ground Cover & Home Ignition Zone & Zone 0 (0-5 ft) & Yes \\
        S-02 & Vegetation-Structure Contact & Home Ignition Zone & Vegetation Management & Yes \\
        S-03 & Vertical/Ladder Fuel Continuity & Home Ignition Zone & Vegetation Management & Yes \\
        S-04 & Overall Vegetation Condition & Home Ignition Zone & Vegetation Management & Yes \\
        S-05 & Combustible Outbuilding Present & Home Ignition Zone & Other Fuels & \\
        S-06 & Firewood Pile/Fuel Tank Adjacent to Structure & Home Ignition Zone & Other Fuels & \\
        L-01 & Terrain Slope & Landscape Context & Topography & Yes \\
        L-02 & Adjacent Parcel Condition & Landscape Context & Neighborhood & Yes \\
        L-03 & Power Lines & Landscape Context & Infrastructure & Yes \\
        L-04 & Road Access & Landscape Context & Infrastructure & Yes \\
        \hline
    \end{tabular}
\end{table}

After filtering the JSON dictionary to retain only these 16 critical contributing features, the structured data must be reformatted into a semantic structure that is easily digestible by the CLIP language encoder.

\subsection{Natural Language Conversion}
\label{app:natural_language}

The final step of the data preparation pipeline is a rule-based structured-to-unstructured conversion. To maintain data density and prevent the CLIP encoder from wasting attention on irrelevant information, benign or absent risk states such as a feature evaluated as ``false'' are omitted entirely rather than explicitly described. The remaining confirmed vulnerabilities are concatenated into a cohesive, comma-separated descriptive caption.

Based on the filtered attributes from the example output in Section \ref{app:example_outputs}, the final generated natural language text passed to the CLIP encoder is formulated as follows:

\begin{promptbox}
"wood structure, pitched tile roof, stucco/cement siding to ground contact, non-combustible deck (enclosed underneath), combustible attached fence, flat slope, mowed lawn, green and well-maintained veg, power line encroachment, paved and wide road"
\end{promptbox}

\end{document}